\documentclass[journal]{IEEEtran}
\usepackage{graphicx}
\usepackage{amsmath}
\usepackage{multirow}
\usepackage{booktabs}
\usepackage{url}
\usepackage{subfigure}
\usepackage{xcolor}
\usepackage{slashbox}
\usepackage{array}
\usepackage{bm}
\usepackage{algorithm}
\usepackage{algpseudocode}
\usepackage{amsmath}

\usepackage{amsfonts,amssymb}
\usepackage{mathrsfs}
\usepackage{booktabs}
\usepackage{graphicx}
\usepackage{epstopdf}
\usepackage{cite}
\usepackage[colorlinks,
            linkcolor=red,
            anchorcolor=blue,
            citecolor=green,
            backref=page
            ]{hyperref}
\usepackage{cleveref}

\usepackage{fancyhdr}

\begin{document}

\title{Co-Saliency Detection with Co-Attention Fully Convolutional Network}

\author{Guangshuai Gao,
        Wenting Zhao,
        Qingjie Liu $^{*}, \emph{Member}, \emph{IEEE}$
        and Yunhong Wang, \emph{Fellow}, \emph{IEEE}
\thanks{This work is partly funded by National Key R$\&$D Program of China (No. 2018YFB1701600) and National Natural Science Foundation of China (No.U1804157 and No.41871283). (Corresponding author: Qingjie Liu)

Guangshuai Gao, Qingjie Liu and Yunhong Wang are with the State Key Laboratory of Virtual Reality Technology and Systems, Beihang University,
Xueyuan Road, Haidian District, Beijing, 100191, China and Hangzhou Innovation Institute, Beihang University, Hangzhou, 310051, China. E-mail:\{gaoguangshuai1990, qingjie.liu, yhwang\}@buaa.edu.cn.

Wenting Zhao is with AI Lab of China Merchants Bank, Shenzhen, China. E-mail: wtzhao@buaa.edu.cn.}}

\markboth{IEEE Transactions on Circuits and Systems for Video Technology}%
{Shell \MakeLowercase{\textit{et al.}}: Bare Demo of IEEEtran.cls for IEEE Journals}

\maketitle

\thispagestyle{fancy}
\fancyhead{}
\lhead{}
\lfoot{Copyright \copyright 20xx IEEE. Personal use of this material is permitted. However, permission to use this material for any other purposes must be obtained from the IEEE by sending an email to pubs-permissions@ieee.org.}
\cfoot{}
\rfoot{}

\begin{abstract}

Co-saliency detection aims to detect common salient objects from a group of relevant images. Some attempts have been made with the Fully Convolutional Network (FCN) framework and achieve satisfactory detection results. However, due to stacking convolution layers and pooling operation, the boundary details tend to be lost. In addition, existing models often utilize the extracted features without discrimination, leading to redundancy in representation since actually not all features are helpful to the final prediction and some even bring distraction. In this paper, we propose a co-attention module embedded FCN framework, called as Co-Attention FCN (CA-FCN). Specifically, the co-attention module is plugged into the high-level convolution layers of FCN, which can assign larger attention weights on the common salient objects and smaller ones on the background and uncommon distractors to boost final detection performance. Extensive experiments on three popular co-saliency benchmark datasets demonstrate the superiority of the proposed CA-FCN, which outperforms state-of-the-arts in most cases. Besides, the effectiveness of our new co-attention module is also validated with ablation studies.
\end{abstract}

\IEEEpeerreviewmaketitle

\section{Introduction}
\label{sec:intro}
\IEEEPARstart{W}{ith} the popularity of smart phones and social media, plentiful images which can be organized into groups containing similar objects or events are more easily acquired. Among the various interesting questions that have been brought, one is ``where do people look when comparing images?" Jacobs et al.~\cite{jacobs2010cosaliency} first raised this question and defined a co-saliency detection task, i.e. to identify and segment common and salient foreground objects from a pair or group of related images. Compared with saliency detection over single images that aims to detect salient regions without considering co-occurrence of multiple object instances, co-saliency detection takes a further step to filter out non-common foreground and background, and leaves only the common salient objects across multiple images. This process can benefit many vision applications without requiring to know contents of the images, such as image segmentation/co-segmentation~\cite{wang2019inferring,lai2019video,chang2011co,wang2017video}, video saliency and segmentation~\cite{wang2019revisiting,wang2015saliency,wang2018saliency}, weakly supervised object detection~\cite{siva2013looking,siva2011weakly}, and stereo saliency detection~\cite{wang2016stereoscopic}.

Existing approaches generally focus on tackling three key problems to achieve co-saliency detection~\cite{zhang2018review}: 1) extracting representative features to characterize salient objects; 2) mining co-saliency cues; 3) designing computational frameworks to generate co-saliency maps.
The feature representations utilized in previous methods include low-level hand-crafted features~\cite{chang2011co,fu2013cluster,tan2013image} (e.g. Gabor, SIFT, or their combinations), mid-level attributes~\cite{fu2013cluster,tan2013image,ge2016co,zhang2016detection,liu2014co,li2015efficient,cong2018co} (i.e. combining predictions from previous saliency or co-saliency detection algorithms), and high-level semantic features~\cite{yao2017revisiting,zhang2016cosaliency,zhang2015co,zhang2015self} from some deep networks.
With the extracted features, intra- and inter-saliency of the input images are modelled, from the perspective of individual images and multiple relevant images respectively. Most early  modelling frameworks of co-saliency detection~\cite{fu2013cluster,li2013co,li2014co,li2014efficient} are designed in a bottom-up manner. They heavily rely on human knowledge for designing hand-crafted metrics to explore the intrinsic patterns underlying the co-salient objects, and suffer poor generalization ability to diverse real-world scenarios. There are also some methods~\cite{cao2014self,chen2014implicit,huang2015saliency,tsai2017segmentation,ye2015co,cao2013saliency,cao2014co} that fuse predictions or mined knowledge from existing (co-)saliency detection algorithms to produce final co-saliency maps. These methods can work well in various scenarios but their performance is largely decided by the existing detection algorithms they adopt. With wide success in a variety of vision tasks, deep learning techniques have also been applied to co-saliency detection~\cite{zhang2015self,zhang2015co,zhang2016detection,zhang2017co,wei2017group,wei2019deep,zhang2019co} which directly learn co-saliency patterns from the given image group and usually offer good performance. But these learning-based models are data-hungry and sometimes hard to converge.

Recently, fully convolution neural networks (FCNs)~\cite{long2015fully}
have motivated several advances in
co-saliency detection.
In~\cite{wei2017group,wei2019deep,zhang2019co}, group-wise deep frameworks are developed based on FCN, which can automatically learn high-level semantic features and model collaborative relationships of feature representations from group-wise and individual images, with competitive and reliable results obtained.
However, FCN-based models may suffer from several disadvantages which severely limit the detection performance: 1) These models extract features over all the pixels without discrimination, which leads to redundancy in representation since not all the features are useful to the prediction of co-saliency, and some even bring distraction to final results. 2) The stacking convolutional layers and pooling operations in FCN-based models decrease the size of feature maps, which result in boundary detail loss.

With these insights, in this paper, we propose to enhance the discriminative power of the model by introducing a Co-attention module to add weights on the image feature maps. The module can adaptively allocate larger weights on the common salient regions, meanwhile smaller on the uncommon backgrounds and distractors. With this module, feature information can be fully utilized, and leading to considerable performance improvements.

Accordingly, we embed a novel co-attention module to the convolutional layers of FCNs, and build an end-to-end deep supervised framework for co-saliency detection (named CA-FCN for brevity) which accepts pairwise input images and outputs co-saliency maps. The feature maps from the last several convolutional layers in CNNs usually contain more semantic information that is beneficial to accurate co-saliency detection. The proposed co-attention module is therefore built on top of these feature maps for better highlighting the co-salient objects while suppressing the uncommon distractors or background.
In particular, our proposed CA-FCN framework is composed of two branches, each being an FCN structure with three key modules: feature extraction module (FME), co-attention module (CAM) and co-saliency map generation module (CSGM). FME is used to extract multi-scale features; CAM is built on feature maps of the last convolution layers to capture more semantic information to help acquire consistent features between co-saliency images. When the feature maps are weighted via co-attention operation, we concatenate the weighted feature maps with the individual image feature to filter out un-obvious part information. Finally, both of the branches recover their co-saliency maps via CSGM by de-convolution. Extensive experiments on challenging datasets demonstrate the superiority of the proposed CA-FCN architecture compared with state-of-the-art methods.

In summary, we make three-fold contributions in this paper:

1. We propose a co-attention module that acts like a mask to allocate larger weights on the co-salient regions and smaller weights to filter out backgrounds and non-common distractors. It  greatly enhances the model's discriminative power to co-saliency information and therefore lifts the final performance.

2.	We build CA-FCN, an end-to-end supervised deep framework based on FCN, which can effectively exploit semantic information through pixel-level classification over input images, and achieve higher sensitiveness to co-saliency information assisted by our novel co-attention module.

3. Extensive experiments demonstrate that our proposed CA-FCN can achieve very good results on three commonly used challenging benchmark datasets, in comparison with state-of-the-art unsupervised and supervised saliency and co-saliency detection methods. Also, the superiority of our co-attention module is well validated through ablation studies.

The remainder of our paper is organized as follows. Section~\ref{sec:literature} reviews the previous saliency/co-saliency detection models as well as related attention models. Section~\ref{sec:method} elaborates on the proposed network architecture. Section~\ref{sec:experiment} provides experiment results in comparison with state-of-the-arts. Finally, a conclusion with future work is given in Section~\ref{sec:conclusion}.
\section{Related works}
\label{sec:literature}
In this section, we briefly discuss some related literatures of saliency detection methods, and introduce some models of co-saliency detection approaches. In addition, some attention-based methods are also mentioned.

\subsection{Single-image Saliency detection}
Humans have the ability to quickly locate objects or regions which attract their attention. In computer vision community, simulating this processing is known as visual attention prediction or visual saliency detection. Generally, existing single-image saliency detection models can be classified into two categories: unsupervised based and supervised based. Unsupervised models are usually bottom-up based, most of which are based on various prior knowledge such as contrast prior~\cite{liu2007learning,cheng2015global}, frequency domain prior~\cite{achanta2009frequency}, saliency transfer~\cite{wang2016correspondence}, background prior~\cite{zhu2014saliency,wang2016background} and compactness prior~\cite{zhou2015salient}, to name a few. Supervised models mainly based on deep learning techniques, which have shown remarkable improvement than unsupervised ones. DVA~\cite{wang2017deep} incorporated multi-level features into a single network. ASNet~\cite{wang2019inferring} learned to detect the salient detection from human fixations. Beyond of the scope of the paper, more detailed introduction of salient detection can be referenced in recent surveys~\cite{borji2015salient,cong2018review}.

\subsection{Co-saliency detection}
Compared with traditional saliency detection which aims at locating the salient objects from a single image, co-saliency detection aims to identify the informative and attractable objects from a group of relevant images with contents, size, categories totally unknown. Therefore this task is more challenging than saliency detection. Most co-saliency detection models can be employed to detect salient objects for single images by simply reducing the number of processed images to one. According to different correspondence capturing strategies, existing co-saliency detection methods can be roughly classified into  bottom-up methods~\cite{fu2013cluster,tan2013image,ge2016co}, fusion-based methods~\cite{cao2014self,chen2014implicit,ye2015co,tsai2017segmentation} and learning-based methods~\cite{wei2017group,min2018deep}. A more detailed survey can be found in~\cite{zhang2018review} and~\cite{cong2018review}.

Bottom-up methods~\cite{fu2013cluster,tan2013image,ge2016co,zhang2016detection,liu2014co,li2015efficient,cong2018co} score each pixel or region in the images by hand-crafted co-saliency cues. Fu et al.~\cite{fu2013cluster} proposed a cluster method by extracting three low-level features including contrast, spatial and correspondence cues. Tan et al.~\cite{tan2013image} presented a self-contained method via computing a super-pixel affine matrix. Ge et al.~\cite{ge2016co} considered the co-saliency detection task as a process of inter- and intra-saliency propagation. Zhang et al.~\cite{zhang2016detection} explored wide and deep information, and combined intra-image contrast, intra-group consistency and inter-group separability in a Bayesian framework. Liu et al.~\cite{liu2014co} put forward a hierarchical segmentation based method by integrating the regional similarity on the basis of fine segmentation and the object prior on top of coarse segmentation, respectively. Li et al.~\cite{li2015efficient} proposed a saliency-guided method which first selects the queries according to saliency maps generated by an existing method, and then refines the detection results at group level using efficient manifold ranking techniques.
These bottom-up methods detect co-saliency from input images with four main steps including pre-processing, feature description, single cue exploration, and multi-cue combination. In the pre-processing step, the input image is segmented into several blocks or superpixels; then the feature vectors are extracted from each block or superpixel to represent the whole image via capturing bottom-up cues; finally, all of these feature vectors are assembled to generate the final co-saliency maps. Though with good performance, they heavily depend on the handcrafted cues and suffer poor generalization ability in the real-world scenarios.

Fusion-based methods~\cite{cao2014self,chen2014implicit,huang2015saliency,tsai2017segmentation,ye2015co,cao2013saliency,cao2014co} mainly excavate useful information from the saliency maps generated from other existing saliency detection or co-saliency detection approaches and then fuse them to generate the final co-saliency maps. Cao et al.~\cite{cao2014self} combined multiple saliency maps via a low-rank matrix recovery technique. Chen et al.~\cite{chen2014implicit} extended the saliency detection task to co-saliency detection in an implicit rank-sparsity decomposition manner. Huang et al.~\cite{huang2015saliency} fused multi-scale saliency maps by utilizing low-rank recovery and adopting a GMM-based co-saliency prior.
Tsai et al.~\cite{tsai2017segmentation} harnessed the object-aware segmentation evidence and region-wise consensus to address co-saliency detection and co-segmentation jointly. Ye et al.~\cite{ye2015co} first generated exemplar saliency maps with single-image saliency detection models, then performed local and global recovery and exploited border connectivity to obtain region-level co-saliency maps, and finally used the foci of attention area based pixel-level saliency derivation to generate pixel-level co-saliency maps.
Fusion-based approaches often get better detection performance than bottom-up ones since they inherit the advantages of individual (co-)saliency detection methods. On the other hand, their performance is largely dependent on those methods and may be suboptimal if most individual methods only offer inaccurate detection results.

Learning-based methods~\cite{wei2017group,min2018deep,zhang2017co,zhang2016cosaliency,han2018a,cong2019iterative,wei2019deep,zhang2019co,song2019easy} have attracted increasing attention since machine learning is much superior and achieves great success for co-saliency detection. Wei et al.~\cite{wei2017group} extracted co-saliency correspondence with an end-to-end fully convolution network architecture. Li et al.~\cite{min2018deep} integrated global and local detail information, and extracted multi-stage features while preserving pixel-level detail information. Zhang et al~\cite{zhang2017co} integrated both multiple instance learning (MIL) and self-paced learning (SPL) into a unified learning framework, with MIL to measure intra-image contrast and inter-image consistency while SPL to alleviate data ambiguity in complex scenarios. Zhang et al.~\cite{zhang2016cosaliency} combined intra-saliency prior transfer and deep inter-saliency mining for co-saliency detection. Han et al.~\cite{han2018a} presented a unified metric learning based framework to jointly learn discriminative feature representation and co-salient object detector. Zhang et al.~\cite{zhang2019co} put forward a mask-guided FCN framework for co-saliency detection.
Learning-base methods usually address co-saliency detection as a classification task for each pixel/region, and make inference from input images by learning a set of training parameters automatically. Our work integrates the feature learning and co-saliency detection into a unified end-to-end deep supervised framework, which also belongs to a learning-based method.

\subsection{Attention models}
The attention mechanism is prominent for its ability to select discriminative features, and has been applied to many computer vision tasks, such as saliency detection~\cite{zhang2018progressive,zhao2019pyramid,ji2018salient}, image captioning~\cite{chen2017sca}, image classification~\cite{hu2018squeeze}, semantic segmentation~\cite{ren2017end}, image deblurring~\cite{qian2018attentive}, and visual pose estimation~\cite{chu2017multi}.

Vaswani et al.~\cite{vaswani2017attention} proposed the first self-attention mechanism based method to describe global dependencies of the inputs to solve image translation. Zhang et al.~\cite{zhang2018progressive} proposed a novel attention guided network which selectively integrates multi-level contextual information in a progressive manner. The work~\cite{zhao2019pyramid} extends the self-attention mechanism to scene segmentation, and delicately designs a dual attention network to capture rich feature representation. In addition,~\cite{wang2018non} exploits non-local operation in spatial-temporal dimension for video classification. Zhang et al.~\cite{zhang2018self} applied the self-attention mechanism to a GAN framework for better image generation.
In this work, inspired by the success of non-local operation~\cite{wang2018non}, we design a novel co-attention module to learn discriminative features to achieve better performance of co-saliency detection.

The concept of ``co-attention'' is not first used in this work but previously investigated in~\cite{lu2016hierarchical,nguyen2018improved,xiong2016dynamic,wu2018you,yu2017multi} which apply the co-attention to capture underlying relationships between different modalities. For instance, Yu et al.~\cite{yu2017multi} designed an end-to-end deep architecture to learn image attention and question attention jointly, so that image regions and corresponding segments of documents can be selectively focused through the learned model.

Also, our work may seem somewhat similar as \cite{chen2018semantic,yu2019what,lu2019see} but actually differ apparently from them. Chen et al.~\cite{chen2018semantic} took advantage of the channel attention to capture semantic information for co-segmentation, while our work  concentrates on the correlations between positions of pixels on the feature maps.
Yu et al.~\cite{yu2019what} constructed a self-supervised attention learning architecture to learn a shared joint attention feature representation of cross-view videos; Lu et al.~\cite{lu2019see} presented an unsupervised framework by leveraging a co-attention mechanism to facilitate the correspondence learning for video object segmentation. Compared with them, we adopt full-supervised learning to make best use of label information in the pixel-wise annotations.
Most importantly, the focus of our work is on mining semantic and contextual information of the pixels' locations on feature maps of the co-salient images with a simple and elegant fully-supervised network architecture. Our approach uses a novel co-attention module as a mask to operate over the feature maps generated by convolution layers in the two branches of an FCN network, which can better highlight co-salient objects and suppress backgrounds and uncommon distractors.

\section{Proposed Method}
\label{sec:method}
The proposed end-to-end supervised FCN framework for co-saliency detection, abbreviated as CA-FCN, incorporates a novel co-attention module to enhance the model sensitiveness to co-saliency information.
Structurally, the CA-FCN is composed of three key components: 1) feature extraction module (FEM) for extracting semantic and contextual features on top of VGG16 backbone network (Section~\ref{subsec:feature}); 2) co-attention module (CAM) built on the feature maps generated from convolution layers of both streams (Section~\ref{subsec:co}); 3) co-saliency maps generation module (CSGM) via de-convolutional operation (Section~\ref{subsec:co-saliency}).

\begin{figure*}[!htb]
	\centering
	\includegraphics[width=1.0\textwidth]{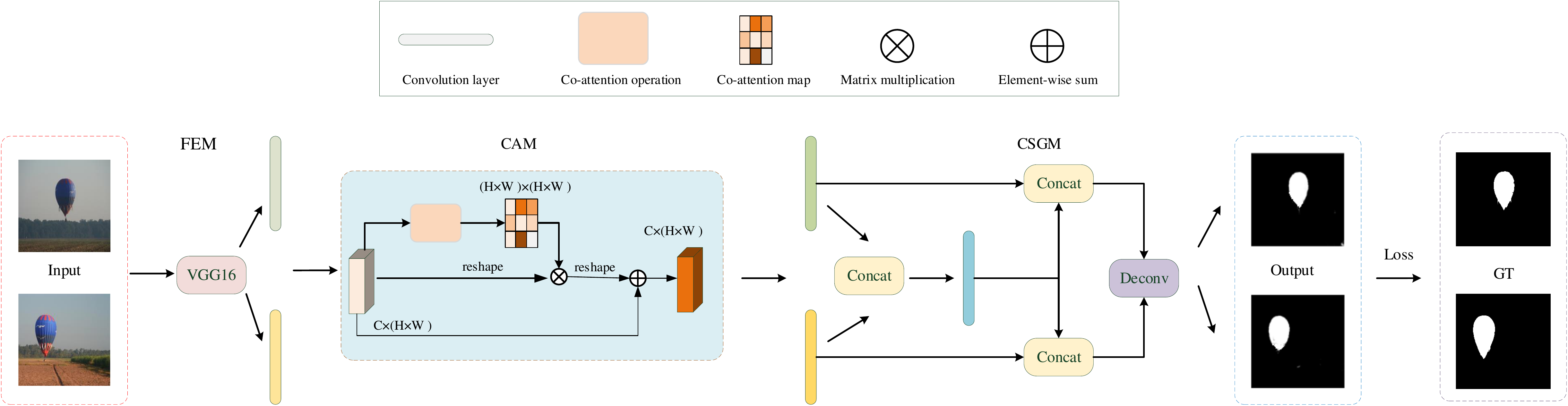}
	\caption{Illustration of proposed CA-FCN architecture. $``\otimes"$ and $``\oplus"$ denote matrix multiplication and element-wise sum, respectively. Note skip connection operation is omitted in the figure.}
	\label{fig:network}
\end{figure*}
\subsection{Architecture overview}
For simplicity, we only consider a pair of images as input to our CA-FCN framework. We can easily modify our model and adapt it to more images, such as done in \cite{wei2017group}. In this way, the model will be too huge. Although trained end to end by pairs of input images in our model, when testing, inspired by \textbf{Group Average Attention} in \cite{chen2018semantic}, our model can be seen as a composition of two parts: an attention-generating module and saliency prediction module.
When processing a group of images $\left[I_{1}, I_{2}, . . I_{\mathrm{n}}\right]$, an attention weight $\alpha_{k}$ will be generated for each image. Each $\alpha_{k}$ corresponds to the disentangled semantic information of each $I_k$. Then the average attention weight $\alpha_{k}$ is used to represent the feature information of each image, so as to further get the common semantic information for the image group. Thus the time complexity is linear time $O(n)$. This demonstrates our method is stable and effectiveness of the strategy.

The overall pipeline is illustrated in Fig.~\ref{fig:network}, and the specific procedure is shown in Algorithm~\ref{alg:Framwork}.
CA-FCN consists of two branches taking  pair-wise inputs and generating corresponding outputs. The two branches share the same parameters (weight vectors and bias) and extract hierarchical features from each image in the input image pair respectively. Similar to~\cite{wei2017group}, at the end of feature extraction, the semantic features of the image pair are concatenated to give the consistency features.
A novel co-attention module, which is our major novelty in this work, is incorporated into the architecture to further boost the performance. Finally, via de-convolution operation, the two branches output their corresponding predicted co-saliency maps.

Our proposed network is built on an FCN structure. We use VGG16~\cite{simonyan2014very} pre-trained on ImageNet~\cite{deng2009imageNet} as backbone to extract contextual and semantic information. Each image is represented by a feature tensor sized $14\times 14\times 512$ after the backbone network. Then three additional convolution layers are applied to encode the features into a more compact form. The size of the encoded features is $7\times 7\times 256$. To accomplish co-saliency detection, the two input images are considered simultaneously. Following~\cite{wei2017group}, we concatenate two features together and apply three convolution layers to extract consistency features used for co-saliency maps prediction. In this process, a novel co-attention module is incorporated on top of the feature maps from the last convolution layers in both branches, to compensate the information loss due to pooling operation. In this way, the model is able to better highlight the common salient objects while suppressing backgrounds and uncommon distractors in the input images. Then, the consistency features are obtained by concatenating weighted features, and then reused to merge the features generated from each individual image. The merging operation is necessary to weaken the salient regions which are not present in all the image group, and enhance the salient regions which are present in all the image group~\cite{wei2017group}. Finally, the de-convolution operation is executed to obtain the co-saliency maps. In addition, to pursue more precise prediction results, inspired by~\cite{ronneberger2015u,long2015fully}, skip connection operation is adopted, combining coarse final layers with finer, earlier layers to provide richer representation information. The skip connection can not only help achieve better results but also reduce the training time of the network. And the specific parameters setting of our proposed network architecture is listed in Table~\ref{Table:structure}, where VGG16 and additional three convolutional layers are included in the stage of feature extraction module (FEM), feature weighted layer are according to co-attention module (CAM), and feature concatenation and co-saliency prediction are included in the stage of co-saliency map generation module (CSGM).

 \begin{algorithm}[htb]
  \caption{Framework of CA-FCN.}
  \label{alg:Framwork}
  \begin{algorithmic}[1]
    \Require
      Image pair (${I_1, I_2}$);
    \Ensure
      Co-saliency maps (${P_1, P_2}$);
    \State Feature extraction module: $(f_1, f_2) = vgg16(I_1, I_2)$;
    \State Co-attention module: $(f_1^w, f_2^w) = CA(f_1, f_2)$;
    \State Concatenating the weighted features $[f_1^w:f_2^w]$ to generate the consistent feature $f_{\text {concat }}$;
    \State Merging $f_{\text {concat }}$ with the features $(f_1^w, f_2^w)$ to generate new features $\left(f_{\text {concat } 1}, f_{\text {concat } 2}\right)$;
    \State Co-saliency map generation: $(P_1, P_2)$ = Deconv$\left(f_{\text {concat } 1}, f_{\text {concat } 2}\right)$;
    \\ End
  \end{algorithmic}
\end{algorithm}

\begin{table}[!htb]
	\centering
	\scriptsize
	\caption{Parameters setting of our network architecture. Since the two branches share the same parameters, we only list the parameters of one of the branches. }	
	\label{Table:structure}	
	\begin{tabular}{lllll}
		\toprule
		& Layer & \ Filter  & Stride &  Output size\\
		\midrule
        \multirow{1}{*}{Input}
        &-- &--&--&$224\times224\times3$\\
        \hline
        \multirow{1}{*} {VGG16}
        &-- &-- &-- &$7\times7\times512$\\
        \hline
		\multirow{3}{*}{Additonal Conv}
		&conv 1&$7 \times 7$& 1 &$7\times7\times1024$\\
		&conv 2&$1 \times 1$& 1 &$7\times7\times1024$\\
		&conv 3& $ 1\times1$ & 1 &$7\times7\times256$\\
		\hline
        \multirow{1}{*} {Feature weighted}
        &-- &-- &-- &$7\times7\times512$\\
        \hline
		\multirow{3}{*}{Feature concatenation}
		& concat.&-&-&$7\times7\times512$\\
		& conv 4&$ 7\times 7$& 1 &$7\times7\times256$\\
		& conv 5&$ 1 \times 1$& 1 &$7\times7\times256$\\
		\hline
		\multirow{6}{*}{Saliency prediction}
		& concat.&-&-&$7\times7\times512$\\
        & conv 6&$ 1 \times 1$& 1 &$7\times7\times256$\\
		& deconv 7&$4 \times 4$& 2 &$14\times14\times512$\\
		& addition &-&-&$14\times14\times512$\\
		& deconv 8&$4 \times 4$& 2 &$28\times28\times256$\\
		& addition &-&-&$28\times28\times256$\\
		& deconv 9&$16 \times 16$& 8&$224\times224\times1$\\

        \hline
        \multirow{1}{*}{Output}
        & co-saliency maps &-&-&$224\times224\times1$\\

		\bottomrule
	\end{tabular}
\end{table}

\subsection{Feature extraction module (FEM)}
\label{subsec:feature}
In the feature extraction module, VGG16~\cite{simonyan2014very} with the last three fully connected layers removed is taken as backbone. Specifically, the first 13 convolutional layers are adopted to capture hierarchical contextual and semantic features. The bottom of our network is a stack of convolutional layers. The inputs and outputs of the convolution layers are a set of arrays, also called feature maps, with size of $h \times w \times c$, where $h$, $w$, and $c$ represent the height, width and channel of the feature map, respectively. The output of each convolutional layer is computed by convoluting the feature map with a trainable linear filter then plus a trainable bias. Denote the input feature map as $X$. The output can be computed as
\begin{equation}
\mathrm{Conv_s} (X;W,b) = W*_s X + b
\end{equation}
where $W$ and $b$ indicate the trainable weight and bias, respectively. $*_s$ denotes the convolution operation with the stride $s$. After convoluting, nonlinear operation (such as Sigmoid, ReLU) is applied to improve the representative capability of features. In addition, down-sampling, such as max-pooling, average pooling, is used following the convolutional layers.

\subsection{Co-Attention module (CAM)}
\label{subsec:co}
\begin{figure}[!tb]
	\centering
	\centerline{\includegraphics[width=0.5\textwidth]{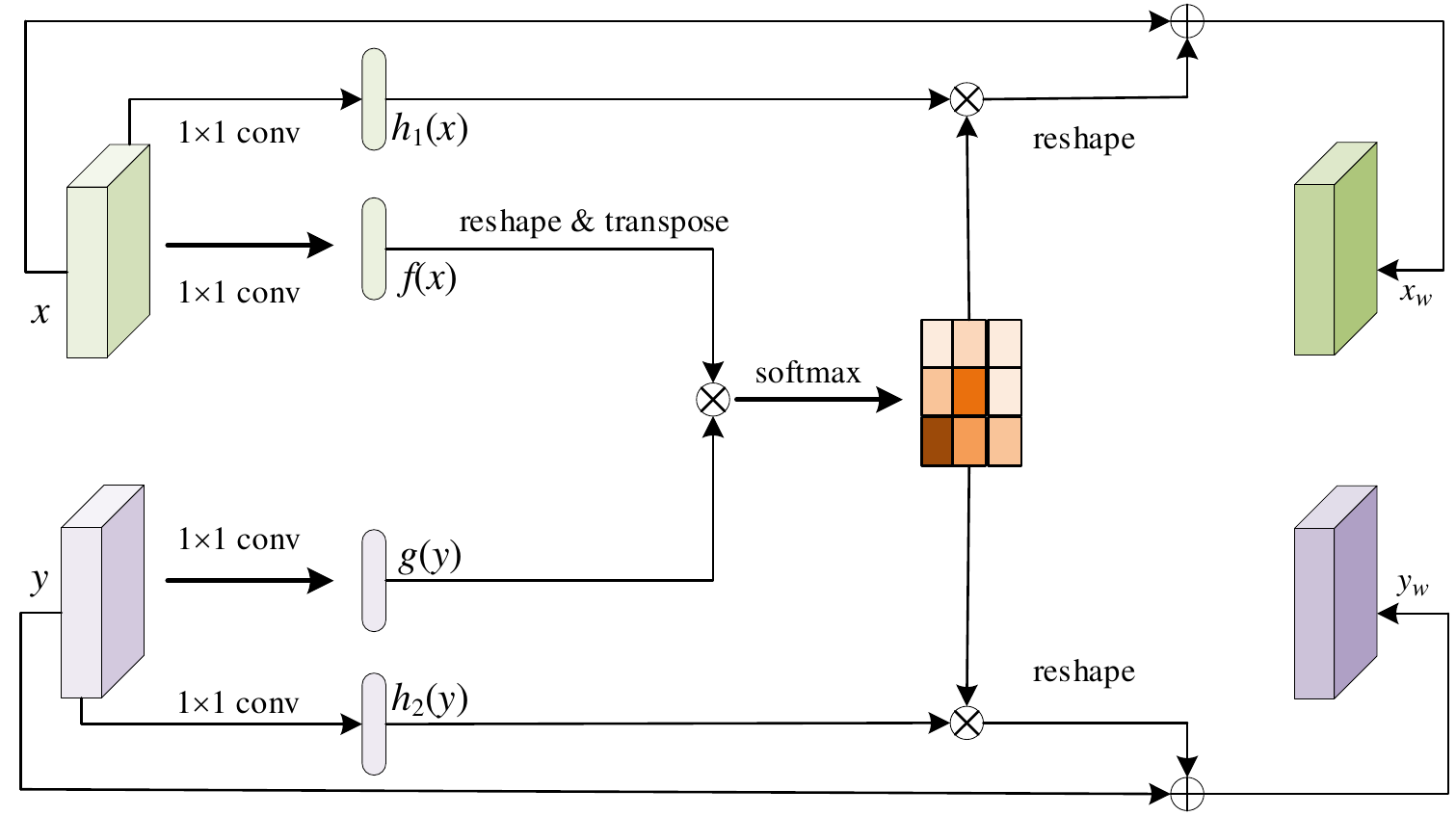}}
	\caption{Structure of proposed co-attention module. $``\otimes"$ and $``\oplus"$ denote matrix multiplication and element-wise sum, respectively. The softmax operation is performed on each row.}
	\label{fig:co-attention}
\end{figure}

The feature maps in the last convolution layers of FME contain abundant contextual information which delivers essential semantic cues and is helpful for the subsequent co-saliency prediction. However, due to multiple stacking convolution layers and pooling operation, the resolution of feature maps is very low, leading to information loss of boundary details. Given an image pair, we expect that the co-saliency detection model can make the common objects highlighted and the background and uncommon distractions weakened. To achieve this goal, we propose to apply the widely used attention mechanism to lean to assign larger weights on the region of interest in the image, and smaller weights on the remainder. We design a co-attention module and incorporate it on top of the feature maps of the last convolution layers of both two branches in our developed framework. The structure of the module is shown in Fig.~\ref{fig:co-attention}.

Concretely, for clarity and without loss of generality, we take $\bm{x} \in \mathbb{R}^{C \times H \times W}, \bm{y} \in \mathbb{R}^{C \times H \times W}$ as the output of the last conv-layers in the two branches, where $C$ denotes the number of channels of the output feature maps, and $H$ and $W$ represent the height and width of the feature maps, respectively.
We transform each of the two features into a new subspace $\textbf{\emph{f}}$, $\textbf{\emph{g}}$ with a $1 \times 1$ convolution operation, respectively. For  convenience of subsequent computation, we first reshape them to $\mathbb{R}^{C \times N}$, where $N=H \times W$ denotes all pixels of the feature map. Afterwards, a matrix multiplication operation between the transpose of $\textbf{\emph{f}}$ and $\textbf{\emph{g}}$ is performed, followed by a softmax layer for computing the co-attention map $\alpha_{i, j}$:
\begin{equation}
\alpha _{i,j}  = \frac{{\exp (s_{i,j} )}}{{\sum\nolimits_{i = 1}^N {\exp (s_{i,j} )} }},\mathrm{where} {\kern 1pt} {\kern 1pt} {\kern 1pt} {\kern 1pt} {\kern 1pt} {\kern 1pt} s_{i,j}  = \bm{f(x_i )^T g\left( {y_j } \right)}
\end{equation}
where \bm{$f(x) = W_f x$}, \bm{$g(y) = W_g y$}, and $\alpha _{i,j}$ denotes the extent of attention the model gives to the $i$-th location when synthesizing the $j$-th region. Then the output of the attention layer is $\bm{o = (o_1 ,o_2 , \cdots o_i , \cdots ,o_j , \cdots ,o_N )} \in\mathbb{R} ^{C \times N}$, where
\begin{equation}
\begin{array}{l}
 \bm{o_j}  = \sum\limits_{i = 1}^N {\alpha _{j,i} } \bm{h(x_i} ),\mathrm{where}{\kern 1pt} {\kern 1pt} {\kern 1pt} {\kern 1pt} {\kern 1pt} {\kern 1pt} {\kern 1pt} \bm{h_1 (x_i )} = \bm{W_{h_1 } x_i};  \\
 \bm{o_i}  = \sum\limits_{j = 1}^N {\alpha _{j,i} } \bm{h(y_j} ),{\kern 1pt} \mathrm{where}{\kern 1pt} {\kern 1pt} {\kern 1pt} {\kern 1pt} {\kern 1pt} {\kern 1pt} \bm{h_2 (y_j )} = \bm{W_{h_2 } y_j }. \\
 \end{array}
 \label{eq:o}
\end{equation}

In the above equations, $\bm{W_f}  \in \mathbb{R}^{\overline C  \times C}$, $\bm{W_g}  \in \mathbb{R}^{\overline C  \times C}$, $\bm{W_{h_1 }}  \in \mathbb{R}^{C \times C}$,
$\bm{W_{h_2 }}  \in \mathbb{R}^{C \times C}$ are the learned weight matrices, which are implemented as $1 \times 1$ convolution. We set $\overline C  = C/8$, as adopted in~\cite{zhang2018self}.

Next, the co-attention layer is added back to the input feature maps by first multiplying a scale parameter. Thus the new co-attention weighted feature maps are given:
\begin{equation}
\begin{array}{l}
 \bm{x_{iw} } = \gamma _1 \bm{o_i}+ \bm{x_i}  \\
 \bm{y_{jw}}  = \gamma _2 \bm{o_j}  + \bm{y_j}  \\
 \end{array}
\end{equation}
where $\gamma _1$ and $\gamma _2$ are learnable parameters, which can be learned through one $1\times1$ convolution layer. They are initialized as 0 and reach optimal values during training. This is to first make the network rely on the cues in the local neighbourhood and then gradually increase the weight on the non-local evidence.

\subsection{Co-saliency maps generation module (CSGM)}
\label{subsec:co-saliency}
Through the co-attention module over the feature maps of the last conv-layers, we obtain the weighted co-attention feature maps. The co-attention module works likes a mask over the original feature maps, assigning larger weights to the pixels on positions of feature maps corresponding to co-salient objects within the input images and lower weights to those in the backgrounds. In this way, our model can make the co-salient objects more highlighted while the background distractors suppressed. Next, we concatenate the weighted feature maps to obtain more informative collaborative high-level features, formally, $\boldsymbol{f}_{\text {concat}}=\left[\boldsymbol{x}_{i w}:\boldsymbol{y}_{j w}\right]$. Nevertheless, it is not sufficient for recovering the predicted co-saliency maps by only depending on the concatenated features. We further concatenate these features, i.e., $\boldsymbol{f}_{\text {concat }}$, with the features extracted from the single image, so as to capture the interaction relationships between pair-wise images and meanwhile retain the unique characteristics of single images. For the final co-saliency maps prediction, we need to up-sample the feature maps to the same resolution with the input images, for which multi-layer deconvolution operations are adopted. In this process, skip connection is used to combine low-level and high-level features to estimate the finer prediction result.

The multi-layer deconvolution operation is formulated as
\begin{equation}
Y = \mathrm{Deconv_s} \left( {\mathrm{Conv_s} \left( {I;\Theta _{conv} } \right);\Theta _{deconv} } \right)
\end{equation}
where $I$ represents the input image pair, and $\mathrm{Conv_s}$, $\mathrm{Deconv_s}$ denote convolution, de-convolution with a stride of $s$. This ensures the output feature maps $Y$ to have the same size with the input image pair. Here, all the parameters of $\Theta$ are learnable.

Finally, we execute a $1 \times 1$ convolution layer on the feature maps $Y$ to generate precisely co-saliency prediction maps $P$ via a sigmoid function. The sigmoid function is used to enforce the entries in the output to have positive values in the range between 0 and 1.

\subsection{Training}
For training, all the parameters $\Theta$ are learnable via minimizing a loss function, which is computed as the error between predicted co-saliency maps and ground-truth images. Since the two branches of our framework have the same network structure, we adopt the same loss function for them. Generally, the salient object only occupies a small fraction of the entire image, in other words, the number of pixels belonging to salient and non-salient regions are usually highly imbalanced. For the imbalanced data cases, an asymmetric weighted loss can greatly improve the performance, as demonstrated in~\cite{mostajabi2015feedforward}. Therefore, similar as \cite{wang2018video}, we adopt a weighted entropy loss function defined as
\begin{equation}
\label{equation:loss}
\mathcal{L}(\mathrm{P},\mathrm{G})=-\frac{1}{MN}\!\sum_{i=1}^{MN}\left((1-\eta) \!g_{i} \log p_{i}+\!\eta\left(1-g_{i}\right) \! \log \left(1-p_{i}\right)\right)
\end{equation}
where $M$ is the number of training image, $N$ means the pixel number of each training image. $p_i$ and $g_i$ indicate the predicted result and ground truth, respectively. $\eta$ denotes the ratio of salient pixels in the ground truths. Empirically, we set $\eta=0.3$.

\section{Experiments}
\label{sec:experiment}

\subsection{Datasets}
To evaluate performance of the proposed method, we use three co-saliency detection benchmark datasets with  pixel-wise ground-truth labels to conduct experiments.

\textbf{iCoseg dataset~\cite{batra2010icoseg}}: The iCoseg dataset consists of 38 groups and a total of 643 images,  with pixel-wise ground-truth annotations. Each group contains 4$\sim$42 images with single or multiple similar objects in diverse sizes, poses and complex backgrounds. Note that only the subset5 which includes 5 images for each group is used in our experiments.

\textbf{MSRC-v2 dataset~\cite{winn2005object}}\footnote{http://jamie.shotton.org/work/data/TextonBoostSplits.zip}: This dataset contains 591 images and 23 object classes with accurate pixel-wise labels. Only 21 object classes are commonly used and the other three classes (void==0, horse==5, mountain==8) are not used since the images in them only contain background and no object or too few training samples. The dataset is widely used for full scene segmentation and object instance segmentation.

\textbf{Cosal2015 dataset~\cite{zhang2016detection}}: This is a new established benchmark dataset, which is more challenging than the other two. It is the largest dataset for co-saliency detection so far. It consists of 50 groups with 2,015 images in total, and each group includes 26$\sim$52 images, involving large variations in pose, size, appearance, and background texture.

\subsection{Implementation details}

Before training and testing, all images and ground truth maps are resized to $224\times 224$ to adapt to the input size of VGG16. The proposed network is trained in an end-to-end manner. In the training phase, the weights of the first five convolutional layers inherit from those of VGG16~\cite{simonyan2014very}, which are pre-trained on ImageNet~\cite{deng2009imageNet}. The parameters of the other layers are initialized randomly. We train the network by minimizing the loss function in Eq.~(\ref{equation:loss}) using stochastic gradient descent (SGD). We set the minimum batch size as 4. The learning rate is set as a value starting from 1e-4 and reduced by a factor of 0.1 in every 50 epochs. 60,000 iterations are needed for convergence. The momentum and weight decay are set to 0.9 and 0.005. The network is deployed with Tensorflow~\cite{abadi2016tensorFlow} platform and trained on a single NVIDIA GeForce GTX Titan 1080 GPU.

For fair comparison, we follow the same protocol in~\cite{siva2013looking,wei2017group}. The training samples are generated from an off-the-shelf image dataset, MSCOCO~\cite{lin2014microsoft}, which contains 9,213 images with pixel-wise ground-truths. Images with the most similar Gist and Lab histogram features are clustered into one group based on their Euclidean distance. In this way, we form our training image pairs, with a total number over 200k.

For testing, we use three groups of testing data, respectively sampled from the three datasets. For iCoseg, we directly use their testing subset built by sampling 5 images per group. For MSRC-v2 and Cosal2015, we randomly sample image pairs from each group and repeat this procedure until all the images over the datasets are covered.

\subsection{Evaluation metrics}
Eight evaluation metrics are used in this paper, including \emph{precision-recall} (P-R) curves, \emph{receiver operating characteristics} (ROC) curves, \emph{F-measure} ($F_\beta$), \emph{mean absolute error} (MAE), \emph{area under ROC curve} (AUC), \emph{average precision} (AP), \emph{weighted F-measure} ($F_\beta ^\omega$) and \emph{structure measure} ($S_\alpha$).

We first perform thresholding with a series of fixed integers from 0 to 255 on co-saliency maps to generate 256 precision and recall pairs, and further plot P-R curves, which is widely used in saliency detection evaluation. The precision and recall are computed as
\begin{equation}
precision = \frac{{\left| {S \cap G} \right|}}{{\left| {S} \right|}},recall = \frac{{\left| {S \cap G} \right|}}{{\left| {G} \right|}}
\end{equation}
where $S$ indicates the set of pixels segmented as the foreground, $G$ denotes the set of pixels labelled as the foreground in the ground truth, and $\left|  \cdot  \right|$ represents the number of elements in a set. Generally, high precision and recall are both required.

ROC curve plots the computation of true positive rate (TPR) and false positive rate (FPR) when binarizing the saliency map with a set of fixed thresholds, with TPR and FPR  defined as
\begin{equation}
TPR = \frac{{\left| {S \cap G} \right|}}{{\left| G \right|}},{\kern 1pt} {\kern 1pt} {\kern 1pt} {\kern 1pt} {\kern 1pt} {\kern 1pt} FPR = \frac{{\left| {S \cap G} \right|}}{{\left| {S \cap G} \right| + \left| {\overline S  + \overline G } \right|}}
\end{equation}
where $\overline S$ and $\overline G$ denote the opposite of the saliency maps \emph{S} and ground-truth \emph{G}, respectively.

Besides, we use $F_\beta$ to balance recall and precision. F-measure can be formulated as
\begin{equation}
F_\beta  {\rm{ =  }}\frac{{\left( {{\rm{1 + }}\beta ^{\rm{2}} } \right) \times precision \times recall}}{{\beta ^{\rm{2}} precision + recall}}
\end{equation}
where $\beta^2$ is set as 0.3, as suggested in~\cite{achanta2009frequency}.

For comprehensive evaluation, MAE is also calculated:
\begin{equation}
MAE = \frac{1}{{W \times H}}\sum\limits_{x = 1}^H {\sum\limits_{y = 1}^W {\left\| {S\left( {x,y} \right) - G(x,y)} \right\|} }
\end{equation}
where \emph{H} and \emph{W} represent the height and width of a saliency map, respectively.

AUC is the area under ROC curve, which distills the information of ROC curve into a single scalar. AP represents the area under the PR-curve while $F_\beta ^\omega$ denotes the weighted F-measure by balancing the \emph{weighted Precision} (a measure of exactness) and \emph{weighted Recall} (a measure of completeness)~\cite{margolin2014evaluate}.

$S_\alpha$ is used to evaluate the spatial structure similarities of saliency maps based on region-aware structure similarity $S_r$ and object-aware structure aware similarity $S_o$, which is defined as
\begin{equation}
S_\alpha   = \alpha *S_r  + (1 - \alpha )S_o
\end{equation}
where $\alpha  = 0.5$ as suggested in~\cite{fan2017structure}.

\begin{table*}[!htb]
	\caption{Quantitative results w.r.t. $\mathbf{F}_{\beta}$, MAE, AUC, AP, $\mathbf{F}_\beta ^\omega$ and $\mathbf{S}_\alpha$. Upper arrow means higher values represent better results, down-arrow denotes lower values indicates better performance. \mbox{\color{red}{\textbf{Red}}}, \mbox{\color{green}{\textbf{green}}} and \mbox{\color{blue}{\textbf{blue}}} indicate the best,  second and third best performance, respectively. ``--'' represents no report.}
	\vspace{-0.3cm}
	\begin{center}
        \resizebox{\textwidth}{!}{
		\begin{tabular}{|l|l|c|c|c|c|c|c||c|c|c|c|c|c||c|c|c|c|c|c|}
			\hline
              \multicolumn{1}{|c|}{\multirow {2}{*}{Methods}} &\multicolumn{1}{c|}{\multirow {2}{*}{Year$\&$Venue}} &\multicolumn{6}{c||}{iCoseg~\cite{batra2010icoseg}} &\multicolumn{6}{c||}{MSRC-v2~\cite{winn2005object}} &\multicolumn{6}{c|}{Cosal2015~\cite{zhang2016detection}} \\
			\cline{3-20}
            ~ &~ &$\mathbf{F}_{\beta}$$\uparrow$ &\textrm{MAE}$\downarrow$ &\textrm{AUC}$\uparrow$ &\textrm{AP}$\uparrow$ &$\mathbf{F}_\beta ^\omega$$\uparrow$ &$\mathbf{S}_\alpha$$\uparrow$ &$\mathbf{F}_{\beta}$$\uparrow$ &\textrm{MAE}$\downarrow$ &\textrm{AUC}$\uparrow$ &\textrm{AP}$\uparrow$ &$\mathbf{F}_\beta ^\omega$$\uparrow$ &$\mathbf{S}_\alpha$$\uparrow$ &$\mathbf{F}_{\beta}$$\uparrow$ &\textrm{MAE}$\downarrow$ &\textrm{AUC}$\uparrow$ &\textrm{AP}$\uparrow$ &$\mathbf{F}_\beta ^\omega$$\uparrow$ &$\mathbf{S}_\alpha$$\uparrow$ \\ \hline
            CBCS~\cite{fu2013cluster} &2013 TIP &0.6885 &0.1922 &0.9294 &0.6124 &0.4757 &0.6374 &0.5206 &0.3677 &0.7030 &0.7130 &0.5361 &0.6178 &0.5130 &0.2268 &0.8261 &0.6083 &0.4254 &0.5870 \\     \hline
            CBCS-s~\cite{fu2013cluster} &2013 TIP &0.6443 &0.1517 &0.9106 &0.5490 &0.4532 &0.6633 &0.5057 &0.2912 &0.6981 &0.6980 &0.5699 &0.7257 &0.4942 &0.1980 &0.8251 &0.5665 &\color{blue}{\textbf{0.4983}} &0.6710 \\    \hline
            CSHS~\cite{liu2014co} &2014 SPL &0.5288 &0.1102 &0.9530 &0.8279 &0.4762 &0.7266 &0.4612 &0.2587 &0.6813 &0.7830 &0.4987 &0.7034 &0.4898 &0.1883 &0.8521 &0.7158 &0.4282 &0.6560 \\ \hline
            IRSD~\cite{chen2014implicit} &2014 ICPR &0.4060 &0.2541 &0.7129 &0.2252 &0.3512 &0.4954 &0.6721 &0.1962 &0.8163 &0.4061 &0.5778 &0.6543 &0.4620 &0.2597 &0.7027 &0.2765 &0.3969 &0.5080 \\ \hline
            IRSD-s~\cite{chen2014implicit} &2014 ICPR &0.3830 &0.2635 &0.6848 &0.2082 &0.3346 &0.4734 &0.6725 &0.1938 &0.8058 &0.3165 &0.5840 &0.6594 &0.4570 &0.2611 &0.6915 &0.2507 &0.4007 &0.5100 \\ \hline
            SACS~\cite{cao2014self} &2014 TIP &\color{blue}{\textbf{0.7099}} &0.1865 &0.9509 &\color{green}{\textbf{0.8582}} &0.5150 &0.7317 &\color{green}{\textbf{0.8530}} &\color{blue}{\textbf{0.1620}} &0.9379 &\color{green}{\textbf{0.9364}} &\color{blue}{\textbf{0.6438}} &0.7772 &\color{red}{\textbf{0.7110}} &0.1904 &\color{blue}{\textbf{0.9331}} &0.8057 &\color{green}{\textbf{0.5582}} &0.7380 \\ \hline
            SACS-s~\cite{cao2014self} &2014 TIP &0.7018 &0.2065 &0.9546 &\color{blue}{\textbf{0.8564}} &0.4676 &0.7176 &\color{green}{\textbf{0.8431}} &0.1916 &\color{green}{\textbf{0.9583}} &0.9132 &0.5728 &0.7698 &0.6750 &0.2285 &0.9207 &0.7812 &0.4791 &0.6900 \\ \hline
            CSCO~\cite{ye2015co} &2015 SPL &0.6790 &0.1286 &\color{red}{\textbf{0.9650}} &0.7872 &\color{green}{\textbf{0.6457}} &\color{red}{\textbf{0.8053}} &0.8429 &\color{green}{\textbf{0.1386}} &\color{blue}{\textbf{0.9539}} &0.8790 &\color{green}{\textbf{0.7010}} &\color{green}{\textbf{0.8110}} &-- &-- &-- &-- &-- &-- \\ \hline
            ESMG~\cite{li2015efficient} &2015 SPL &0.6640 &0.1677 &0.9136 &0.7116 &\color{blue}{\textbf{0.5741}} &0.7093 &0.6245 &-- &0.8228 &0.7834 &-- &0.5804 &0.5114 &-- &0.7691 &0.5133 &-- &0.5446 \\ \hline
            CoDW~\cite{zhang2016detection} &2016 IJCV &0.6830 &0.1930 &\color{green}{\textbf{0.9615}} &0.8480 &0.4886 &0.7467 &0.8039 &0.2116 &0.8979 &0.8420 &0.5665 &0.7281 &0.6490 &0.2462 &\color{green}{\textbf{0.9350}} &\color{blue}{\textbf{0.8134}} &0.4722 &0.7177 \\ \hline
            DIM~\cite{zhang2016cosaliency} &2016 TNNLS &0.6250 &0.1955 &\color{blue}{\textbf{0.9610}} &0.8549 &0.4845 &0.7407 &-- &-- &-- &-- &-- &-- &-- &-- &-- &-- &-- &-- \\ \hline
            GwD~\cite{wei2017group} &2017 IJCAI &0.6983 &\color{blue}{\textbf{0.1018}} &0.9497 &-- &-- &0.7800 &0.5952 &0.2238 &0.6997 &0.8290 &-- &0.7370 &0.6084 &\color{green}{\textbf{0.1434}} &0.8954 &-- &-- &\color{blue}{\textbf{0.7450}} \\ \hline
            SGCS~\cite{tsai2017segmentation} &2017 ICME &0.6357 &0.2067 &0.9472 &0.8264 &0.4668 &0.7055 &0.8217 &0.1919 &0.9517 &\color{green}{\textbf{0.9237}} &0.5734 &0.7617 &\color{blue}{\textbf{0.6940}} &0.2175 &0.9325 &0.8016 &0.4979 &0.7127 \\ \hline
            SP-MIL~\cite{zhang2017co} &2017 TPAMI &0.6385 &0.1941 &0.9563 &0.7122 &0.5319 &0.7575 &0.6231 &0.2085 &0.9391 &0.8386 &0.6200 &0.7750 &-- &-- &-- &-- &-- &-- \\ \hline
            UMLBF~\cite{han2018a} &2018 TCSVT &-- &-- &-- &-- &-- &-- &0.7203 &0.2933 &0.9503 &0.9044 &0.5196 &0.7040 &0.5979 &0.3422 &0.8358 &0.7369 &0.3573 &0.5970 \\ \hline
            UCSG~\cite{hsu2018unsupervised} &2018 ECCV &\color{green}{\textbf{0.7341}} &0.1180 &-- &-- &-- &0.7220 &0.7940 &0.1720 &-- &0.9226 &-- &\color{blue}{\textbf{0.8010}} &0.6920 &\color{blue}{\textbf{0.1590}} &-- &\color{green}{\textbf{0.8149}} &-- &\color{green}{\textbf{0.7506}} \\ \hline
            Gw-FCN~\cite{wei2019deep} &2019 TIP &0.7012 &\color{green}{\textbf{0.1002}} &0.9537 &-- &-- &\color{blue}{\textbf{0.7809}} &0.5993 &0.2238 &0.7018 &0.8290 &-- &-- &0.6105 &\color{red}{\textbf{0.1337}} &0.8997 &-- &-- &-- \\ \hline
            Ours &-- &\color{red}{\textbf{0.7623}} &\color{red}{\textbf{0.0874}} &\color{green}{\textbf{0.9615}} &\color{red}{\textbf{0.8726}} &\color{red}{\textbf{0.7374}} &\color{green}{\textbf{0.7977}} &\color{red}{\textbf{0.8832}} &\color{red}{\textbf{0.0677}} &\color{red}{\textbf{0.9613}} &\color{red}{\textbf{0.9597}} &\color{red}{\textbf{0.8509}} &\color{red}{\textbf{0.8662}} &\color{green}{\textbf{0.6980}} &0.1809 &\color{green}{\textbf{0.9336}} &\color{red}{\textbf{0.8335}} &\color{red}{\textbf{0.5722}} &\color{red}{\textbf{0.7510}} \\ \hline
            \end{tabular}}
    \label{Table:metrics}
	\end{center}
	\vspace{-0.4cm}
\end{table*}

\subsection{Compared Baselines}
We compare our method with representative approaches including CBCS~\cite{fu2013cluster}, CBCS-s~\cite{fu2013cluster}, CoDW~\cite{zhang2016detection}, CSCO~\cite{ye2015co}, CSHS~\cite{liu2014co}, DIM~\cite{zhang2016cosaliency}, ESMG~\cite{li2015efficient}, GwD~\cite{wei2017group}, IRSD~\cite{chen2014implicit}, IRSD-s~\cite{chen2014implicit}, SP-MIL~\cite{zhang2017co}, UMLBF~\cite{han2018a}, SACS~\cite{cao2014self}, SACS-s~\cite{cao2014self}, SGCS~\cite{tsai2017segmentation}, UCSG~\cite{hsu2018unsupervised} and Gw-FCN~\cite{wei2019deep}. Among them, CBCS~\cite{fu2013cluster}, CoDW~\cite{zhang2016detection}, CSCO~\cite{ye2015co}, CSHS~\cite{liu2014co}, ESMG~\cite{li2015efficient}, IRSD~\cite{chen2014implicit}, SP-MIL~\cite{zhang2017co}, SACS~\cite{cao2014self}, SGCS~\cite{tsai2017segmentation} and UCSG~\cite{hsu2018unsupervised} are unsupervised co-saliency methods; DIM~\cite{zhang2016cosaliency}, GwD~\cite{wei2017group}, UMLBF~\cite{han2018a} and Gw-FCN~\cite{wei2019deep} are supervised co-saliency methods; to investigate the influence of interaction information, CBCS-s~\cite{fu2013cluster}, IRSD-s~\cite{chen2014implicit}, SACS-s~\cite{cao2014self}, which are aimed at single image saliency detection, are also used as baselines in our work. When available, we use the publicly released source code with default parameters provided by the authors to reproduce the experiments on our test sets. For some methods without released source code, we either evaluate their metrics on their pre-generated co-saliency maps (CoDW~\cite{zhang2016detection}, CSCO~\cite{ye2015co}, DIM~\cite{zhang2016cosaliency}, ESMG~\cite{li2015efficient}, SP-MIL~\cite{zhang2017co} and UMLBF~\cite{han2018a}), or directly use their statistics reported in their papers (GwD~\cite{wei2017group}, UCSG~\cite{hsu2018unsupervised} and Gw-FCN~\cite{wei2019deep}).

\begin{figure*}[htbp]
	\begin{minipage}[b]{.24\linewidth}
		\centering
		\centerline{\includegraphics[width=6.0cm]{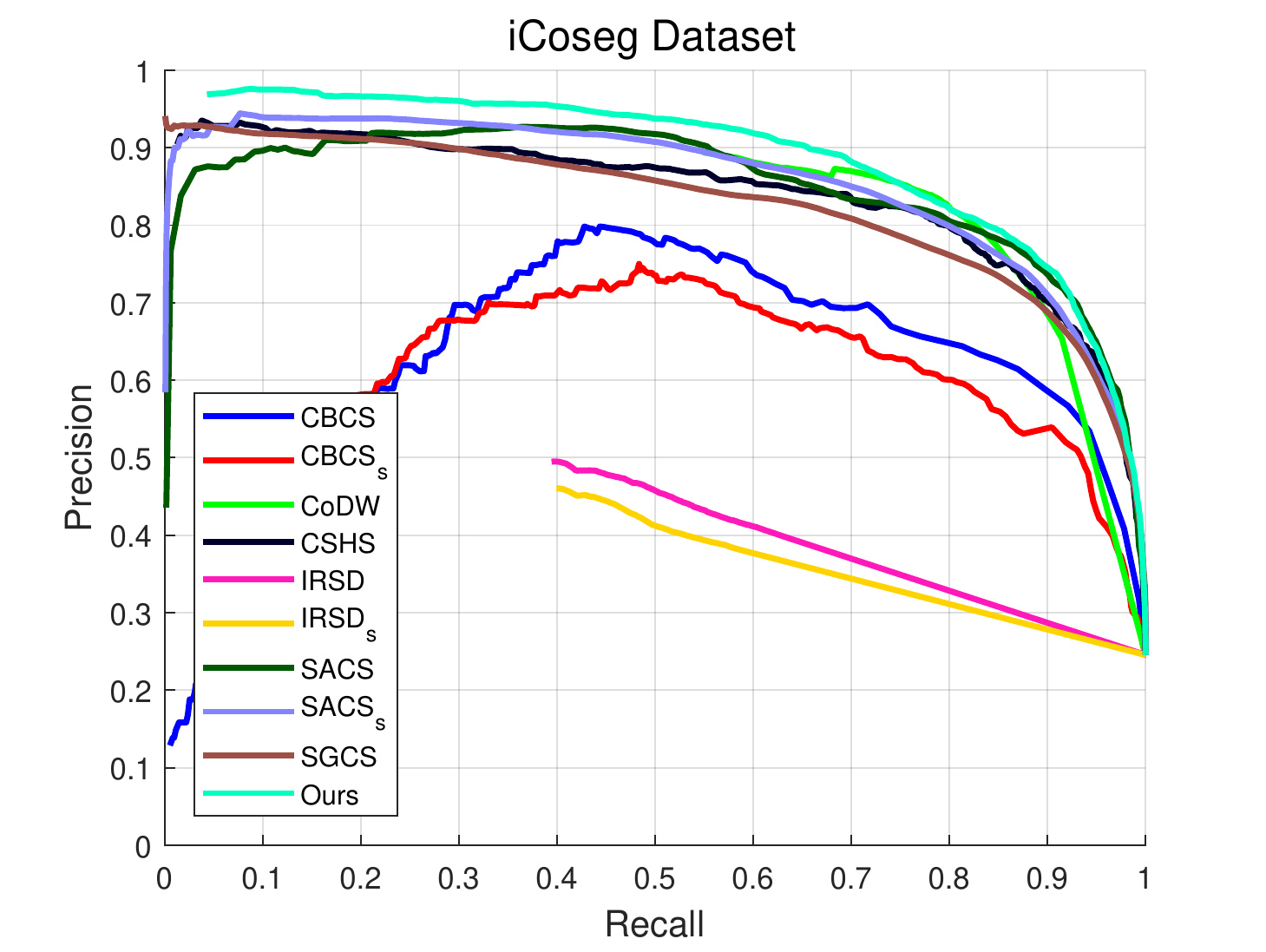}}
	\end{minipage}
	\hfill
	\begin{minipage}[b]{.24\linewidth}
		\centering
		\centerline{\includegraphics[width=6.0cm]{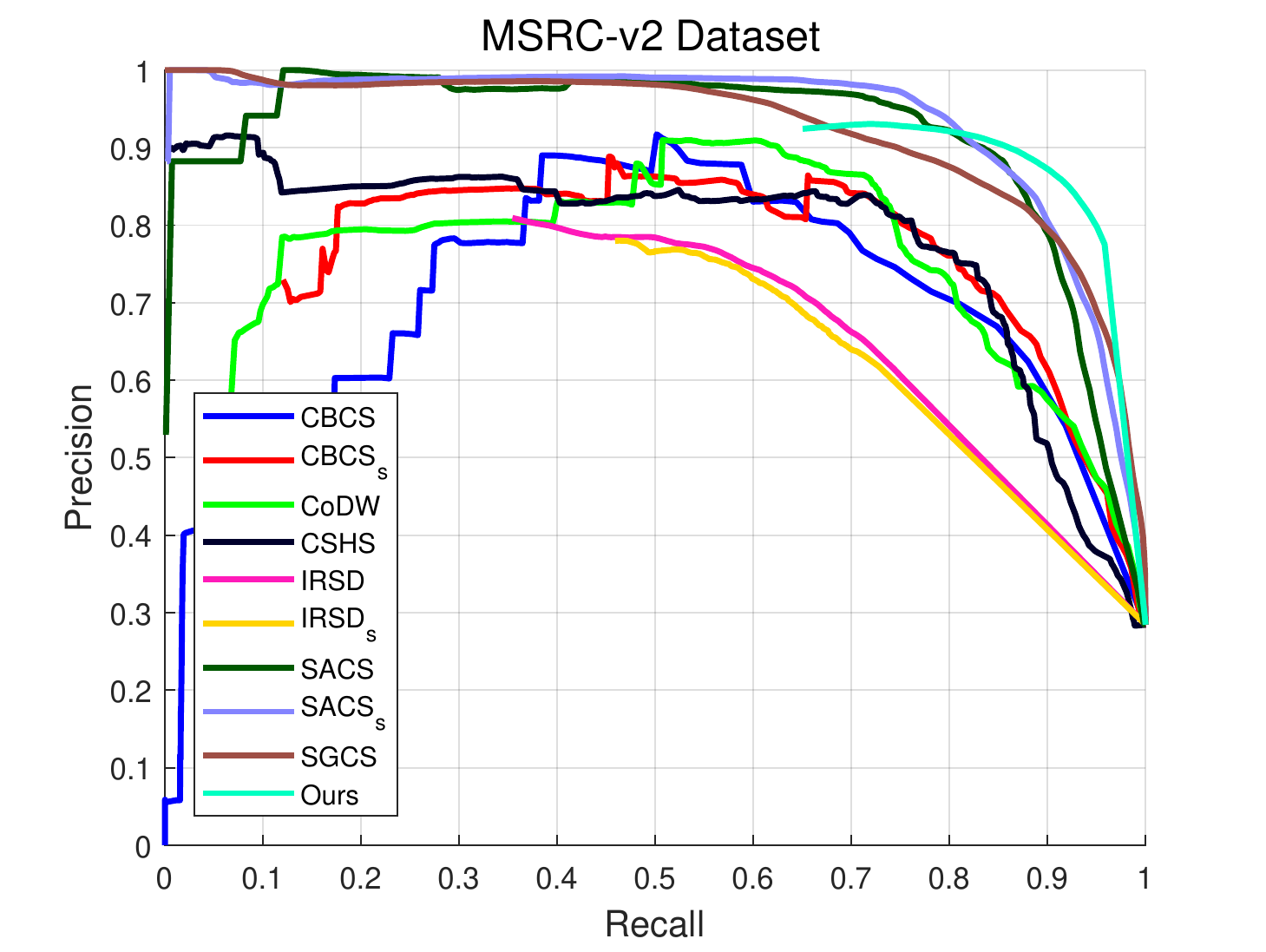}}
	\end{minipage}
	\hfill
	\begin{minipage}[b]{.24\linewidth}
		\centering
		\centerline{\includegraphics[width=6.0cm]{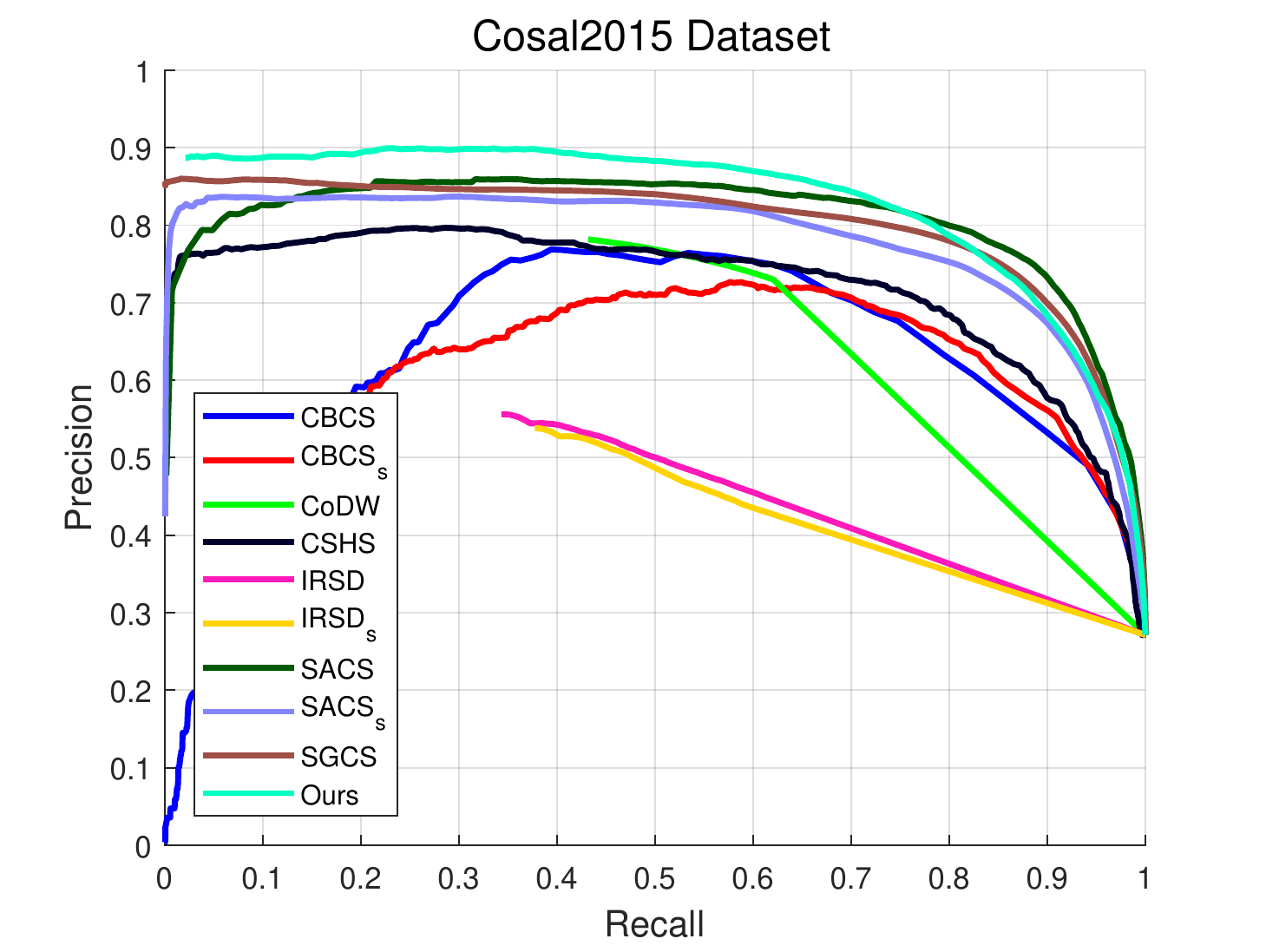}}
	\end{minipage}
	\vspace{-0.3cm}
	\caption{P-R curves of the proposed and other state-of-the-art methods on iCoseg, MSRC-v2, and Cosal2015 datasets.}
	\label{fig:prcurves}
\end{figure*}

\begin{figure*}[htbp]
	\begin{minipage}[b]{.24\linewidth}
		\centering
		\centerline{\includegraphics[width=6.0cm]{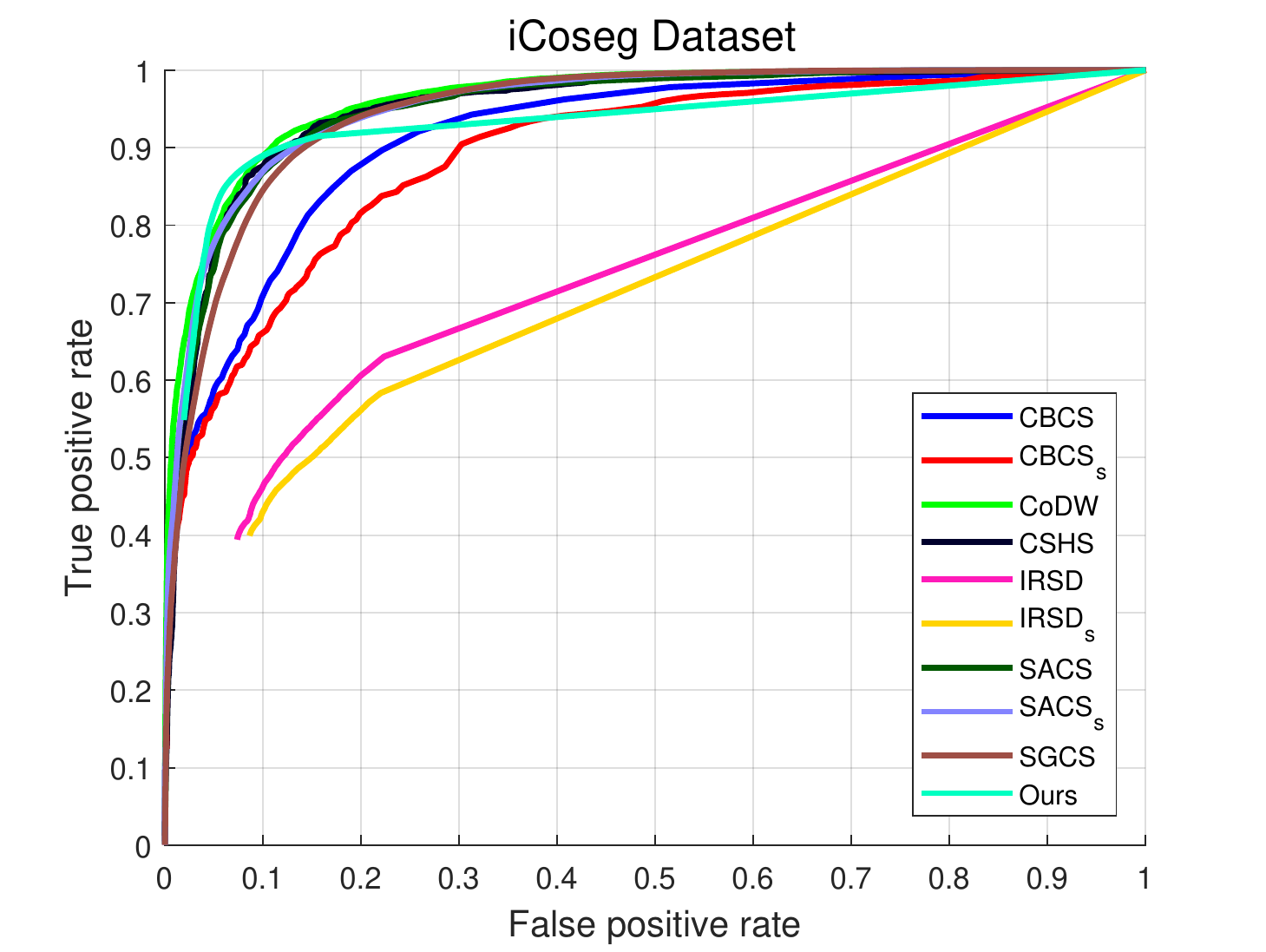}}
	\end{minipage}
	\hfill
	\begin{minipage}[b]{.24\linewidth}
		\centering
		\centerline{\includegraphics[width=6.0cm]{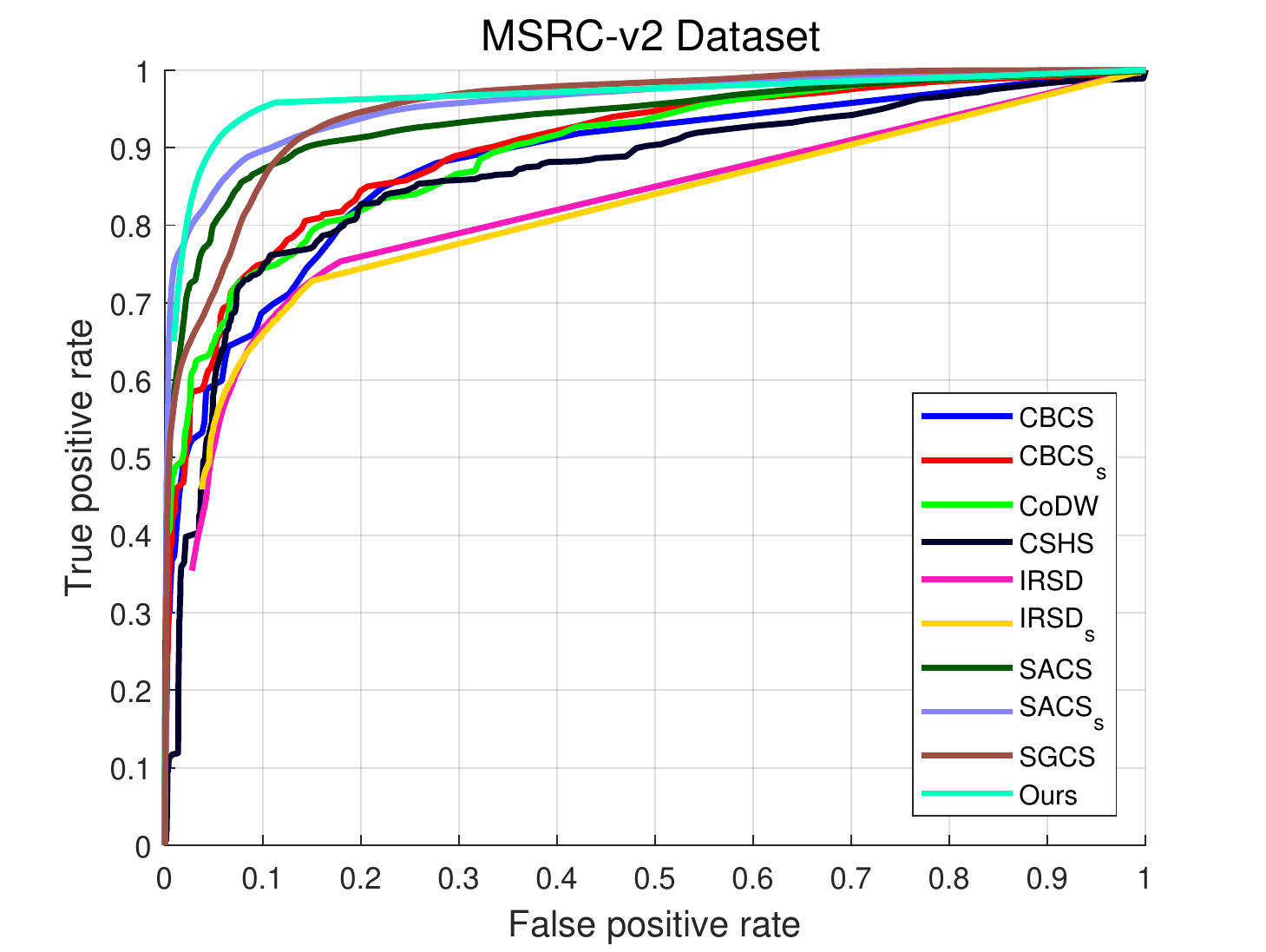}}
	\end{minipage}
	\hfill
	\begin{minipage}[b]{.24\linewidth}
		\centering
		\centerline{\includegraphics[width=6.0cm]{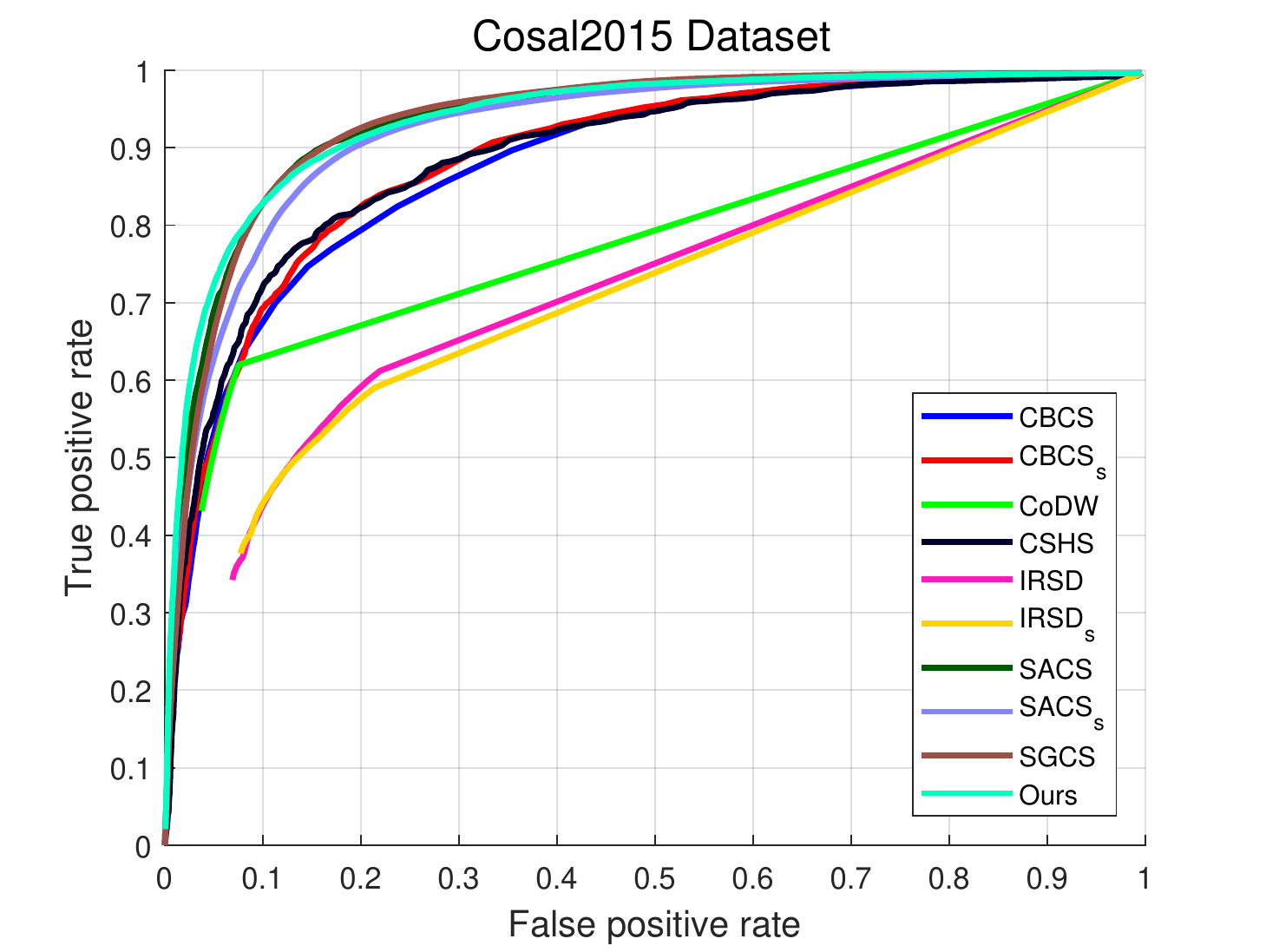}}
	\end{minipage}
	\vspace{-0.3cm}
	\caption{ROC curves of the proposed and other state-of-the-art methods on iCoseg, MSRC-v2, and Cosal2015 datasets.}
	\label{fig:roccurves}
\end{figure*}

\subsection{Comparison with baselines}

The overall performance statistics are provided in Table~\ref{Table:metrics}. PR and ROC curves are shown in Fig.~\ref{fig:prcurves} and Fig.~\ref{fig:roccurves}, respectively.
From Table~\ref{Table:metrics}, it can be seen that our method outperforms other methods by a significant margin in terms of six evaluation metrics  on all the three benchmark datasets. Supervised methods are generally better than unsupervised ones as they are supported by large numbers of object annotations. Even compared with the supervised methods, our approach can yield comparable performance, especially on the MSRC-v2 dataset. Besides, from the PR curves in Fig.~\ref{fig:prcurves} and ROC curves in Fig.~\ref{fig:roccurves}, the proposed method outperforms the state-of-the-arts from an overall perspective.

Specifically, in Table~\ref{Table:metrics}, we can see that our proposed method achieves large improvements in terms of four metrics compared with other baselines on the iCoseg dataset. This depicts the higher precision of our method and lower error between prediction results and ground truths.
Our method is only marginally inferior to CSCO~\cite{ye2015co} regarding AUC and $S_\alpha$ on the iCoseg dataset.
On the MSRC-v2 dataset, our method gives the best performance in terms of all the metrics. On the most challenging dataset Col2015, we can still obtain comparative performance, only slightly lower in some metrics such as AUC compared with CoDW~\cite{zhang2016detection}, which involves many stages and refinement post-processing operations. From an overall perspective, our method performs very favorably on the three datasets at a holistic level.

Furthermore, from Fig.~\ref{fig:prcurves} and Fig.~\ref{fig:roccurves},  we
make the following observations. First, with the same recall, the precision of our method is higher than other competing methods, meaning lower false alarm of our method with the same true positives. Second, with the same precision, our recall is also higher than other methods, which indicates more real co-salient objects can be detected by our method with the same false alarm rate. Third, our method is the highest in terms of the ROC curves, which demonstrates the accuracy of our method is the best among all the methods.

\begin{figure*}[htbp]
\centering
	\centerline{\includegraphics[width=1.0\textwidth]{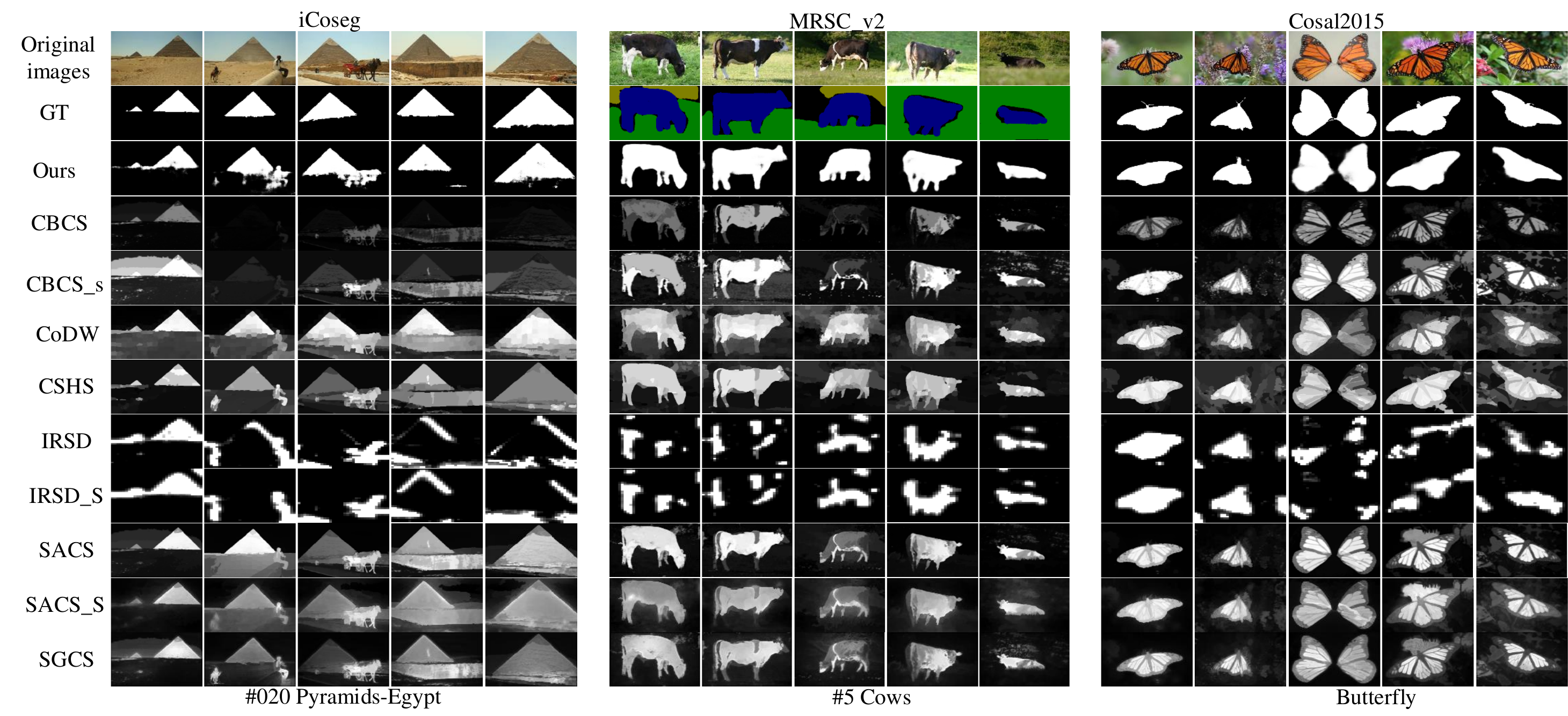}}
	\caption{Visual comparison between the proposed method and the other representative methods on three benchmark datasets.}
	\label{fig:results}
\end{figure*}

We also visualize the co-saliency maps generated by our method and compared methods on the three benchmark datasets in Fig.~\ref{fig:results}. We select one image group from each dataset: Pyramids-Egypt from iCoseg dataset, Cows from MSRC-v2 dataset, and  butterflies from Cosal2015 dataset. As can be seen, our method can better capture the common salient objects. Generally, despite the large variations of testing data, the proposed method can give the best performance on all the three benchmark datasets, which can preserve clearer boundaries between salient regions and backgrounds.

Especially, from Fig.~\ref{fig:results}, we can see in cases where the common objects in relevant images have similar texture or appearance with the background, compared methods tend to perform unsatisfactorily. For example, the fourth and fifth column show the images from Pyramids-Egypt, where the platform and ground have a similar sandy yellow color with the common objects pyramids. CBCS~\cite{fu2013cluster} misses the objects possibly because it is a cluster-based method, and some methods such as SACS~\cite{cao2014self} and SGCS~\cite{tsai2017segmentation}, misclassify backgrounds into objects.
Comparatively, our method can abstract the common objects and highlight them while suppressing the backgrounds.

The cases with large pixel values in backgrounds may cause confusion in the final co-saliency maps. See the second and third columns of cows on MSRC\_v2 dataset. The backgrounds contain grass, which affect the detection of some compared methods such as CSHS~\cite{liu2014co}, CoDW~\cite{zhang2016detection}, SACS~\cite{cao2014self} and SGCS~\cite{tsai2017segmentation}. Comparatively, our method can better separate common objects and backgrounds with clearer boundaries.

The Col2015 dataset is very challenging due to diverse object shapes, colors and very complex scene clutters. For example, on the images from the butterfly set in this dataset, especially the second, fourth and fifth column in Fig.~\ref{fig:results}, we find some methods produce blurred results, such as CoDW~\cite{zhang2016detection}, or incomplete results, such as IRSD~\cite{chen2014implicit}, in the co-saliency maps. In contrast, the results generated by our method show better clearness and completeness, meaning our detections are more accurate and closer to the ground truths.

To sum up, compared with the unsupervised co-saliency detection methods~\cite{fu2013cluster, zhang2016detection,ye2015co,liu2014co,li2015efficient,chen2014implicit,cao2014self,tsai2017segmentation,zhang2017co}, our proposed method can find more complete co-salient objects; compared with the supervised co-saliency detection methods~\cite{zhang2016cosaliency,han2018a}, it can retain more clearer backgrounds; compared with the methods for single image saliency detection, it can make the co-salient objects more highlighted and the backgrounds more suppressed. These results and analyses well speak for the effectiveness of our method.

\subsection{Ablation study}
To gain insight of our proposed CA-FCN model, we conduct ablation studies to investigate the effectiveness of each component in it.

\begin{figure}[htbp]
\centering
	\centerline{\includegraphics[width=0.5\textwidth]{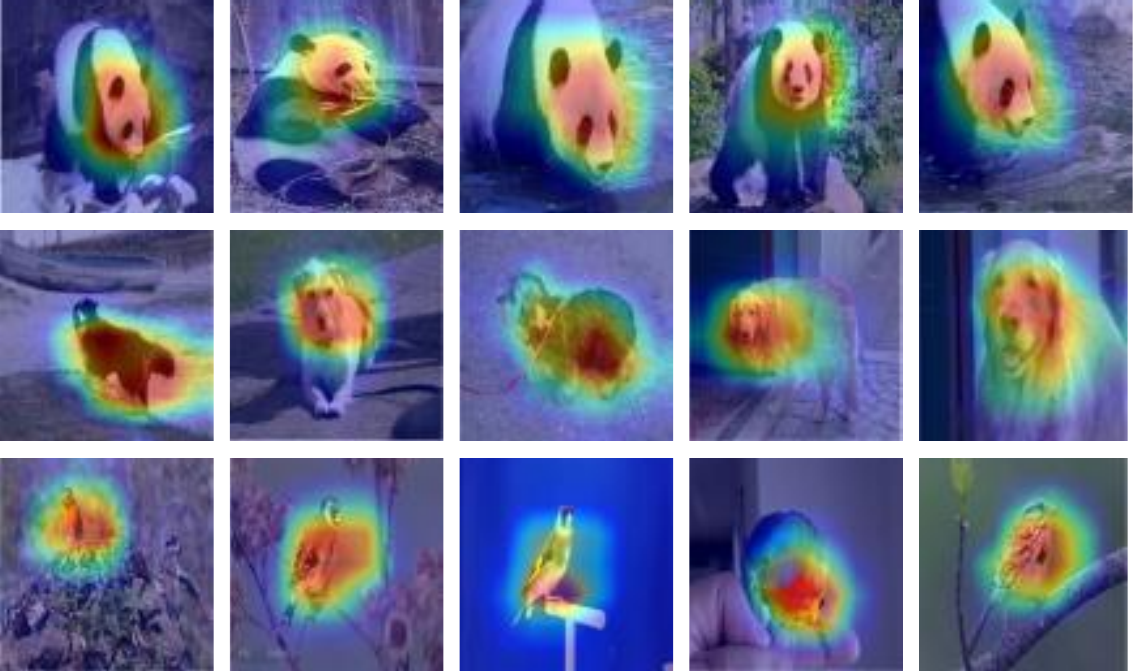}}
	\caption{Visualization of co-attention maps. From top to bottom rows, the examples of co-attention maps from the iCoseg, MSRC-v2, and Cosal2015 dataset are shown respectively. The co-salient objects are denoted with red color. The redder color of common objects means more attention paid to the corresponding positions. The co-attention weights decrease with the colors changing from red to blue.}
	\label{fig:visualization}
\end{figure}

\textbf{The efficiency of co-attention module.} To validate the contribution of our proposed co-attention mechanism, we first visualize the co-attention maps. The heat-maps of co-attention maps for image groups selected from the three benchmark datasets are shown in Fig.~\ref{fig:visualization}. We can see that the co-attention mechanism can focus on the co-salient objects. The redder color of the common objects means more attention paid to the positions, i.e., co-salient objects.

\begin{figure*}[htbp]
	\begin{minipage}[b]{.25\linewidth}
		\centering
		\centerline{\includegraphics[width=6.0cm]{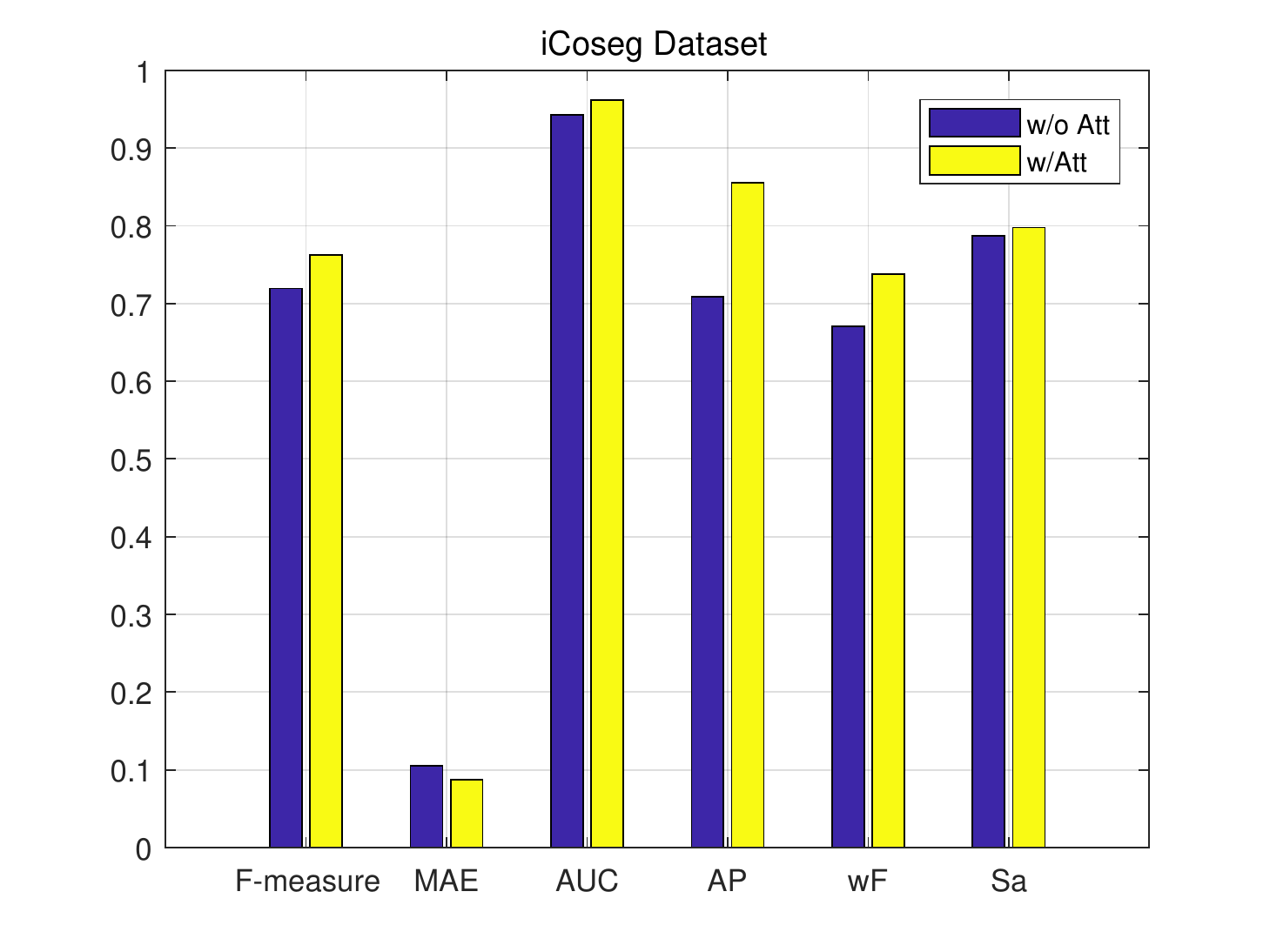}}
	\end{minipage}
	\hfill
	\begin{minipage}[b]{.25\linewidth}
		\centering
		\centerline{\includegraphics[width=6.0cm]{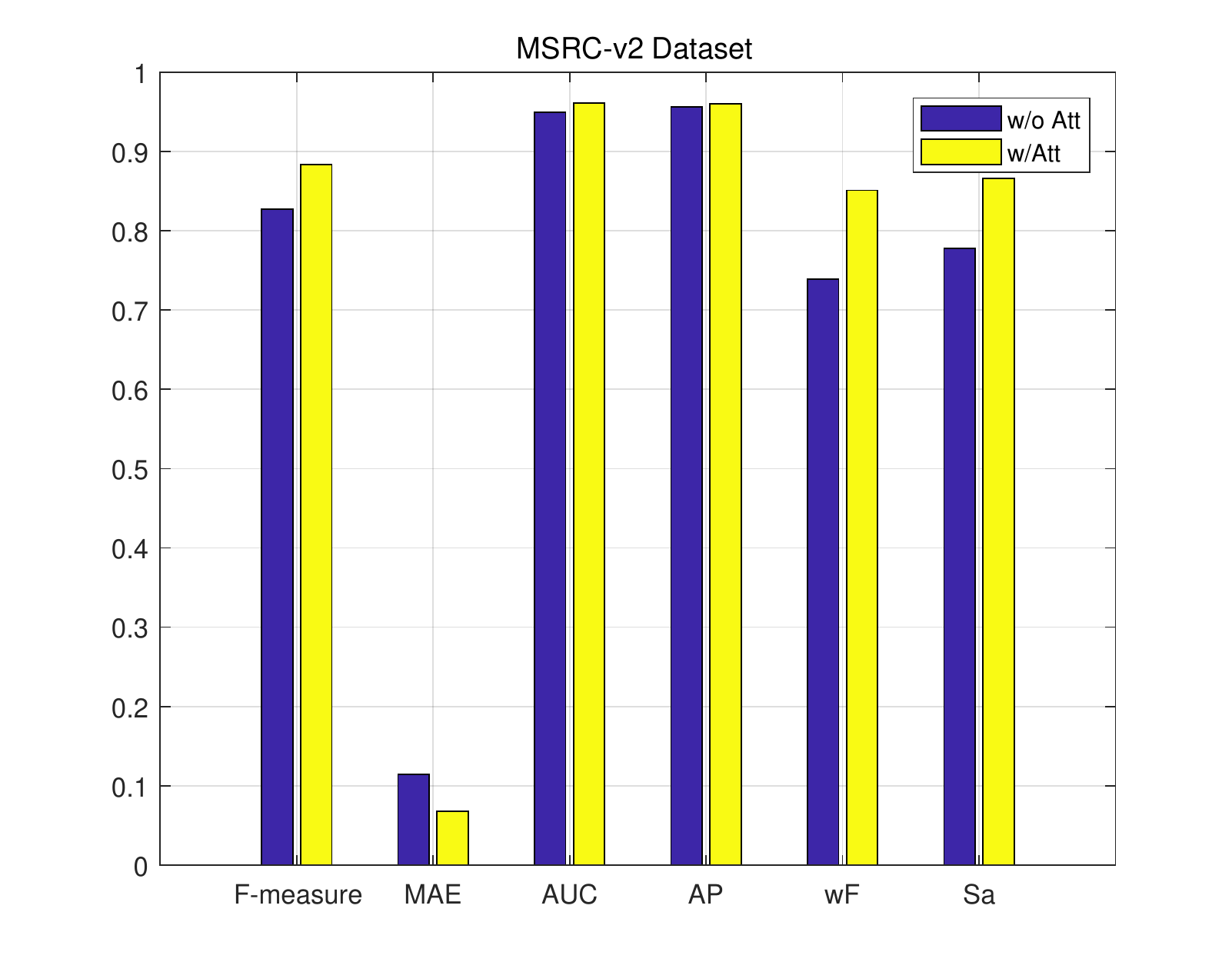}}
	\end{minipage}
	\hfill
	\begin{minipage}[b]{.25\linewidth}
		\centering
		\centerline{\includegraphics[width=6.0cm]{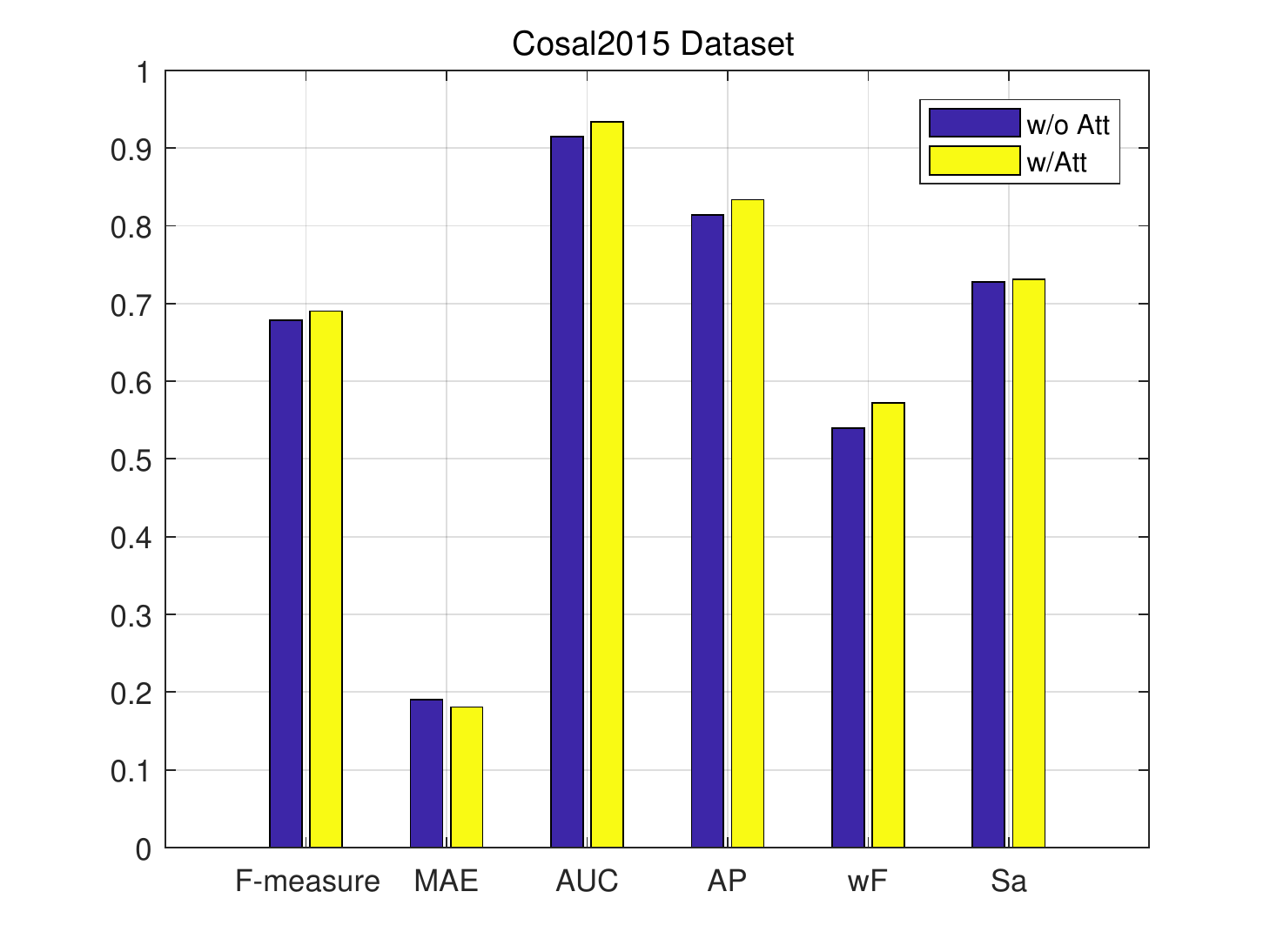}}
	\end{minipage}
	\vspace{-0.3cm}
	\caption{Ablation studies of six evaluation metrics on three benchmarks. $\mathbf{F}_{\beta}$, MAE, AUC, AP, $\mathbf{F}_\beta ^\omega$ and $\mathbf{S}_\alpha$ on each dataset are given, which well show the effectiveness of co-attention mechanism.}
	\label{fig:ablation}
\end{figure*}

Fig.~\ref{fig:ablation} reports the comparison of our CA-FCN without and with co-attention module in different metrics on three benchmark datasets. From the figure we can find the performance is further improved by incorporating the co-attention mechanism into our framework, which well proves the effectiveness of the co-attention module.

\begin{figure*}[htbp]
\centering
	\centerline{\includegraphics[width=1.0\textwidth]{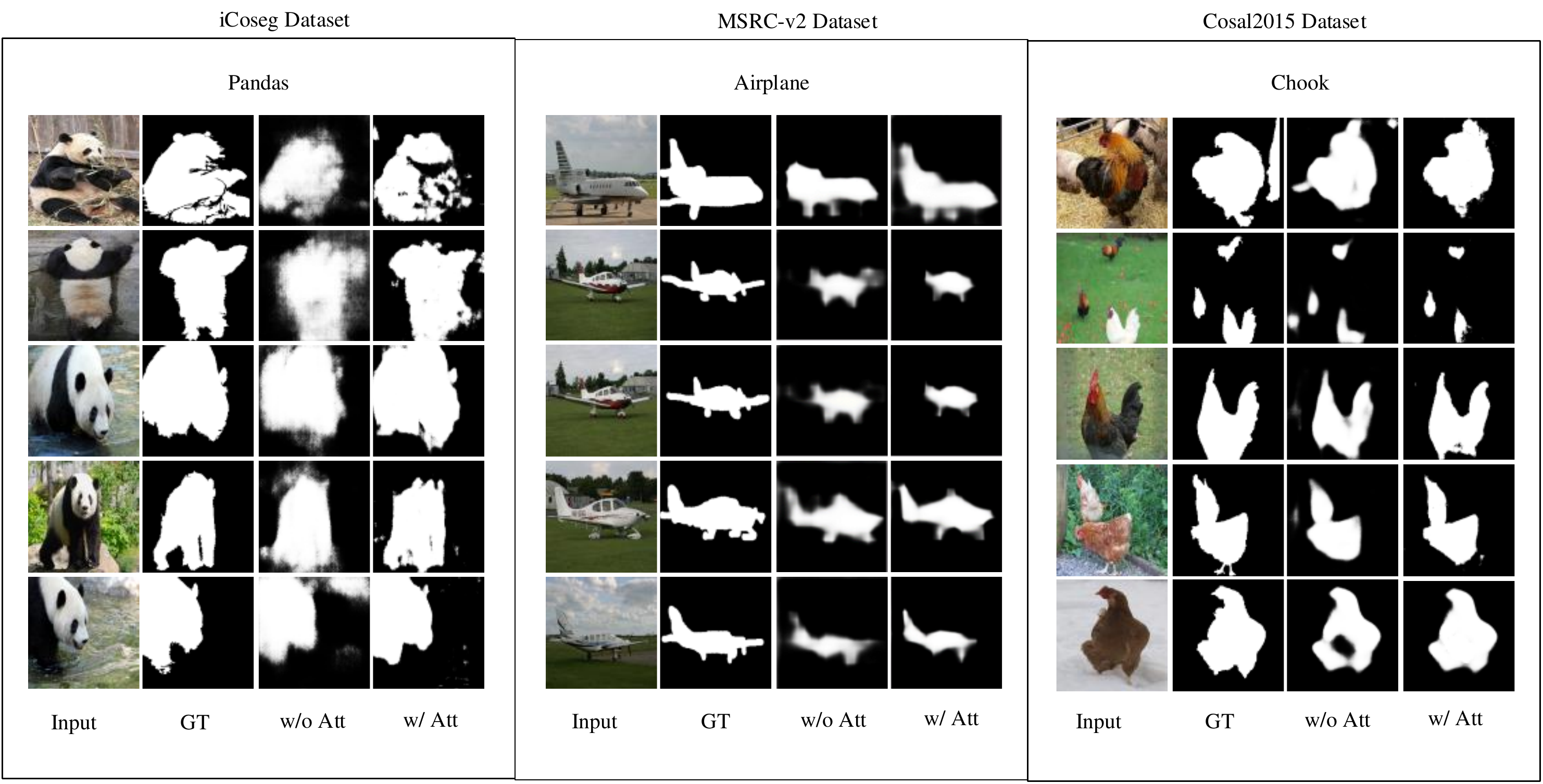}}
	\caption{Comparison of co-saliency maps generated by ours w/o att and w/att methods on three benchmark datasets. w/o att and w/att represent without and with attention module. First column: input image groups;  second column: ground-truth images; third and fourth columns: w/o and w/att images, respectively.}
	\label{fig:comparison}
\end{figure*}

Fig.~\ref{fig:comparison} shows some visual examples of co-saliency maps for ablation studies. We can see  without the attention mechanism, the results may be fuzzy (such as the first and second row in the examples of pandas set), distracted by cluttered background (such as the second row in the examples of airplane set and chook set), or incomplete (such as the fifth row of airplane set and chook set). However, when incorporating the co-attention mechanism into the framework, the common objects are more highlighted and the backgrounds are more suppressed. Overall, by merging the co-attention mechanism, we can improve the performance both in quantitative and qualitative results.

\begin{figure}[htbp]
\centering
	\centerline{\includegraphics[width=0.5\textwidth]{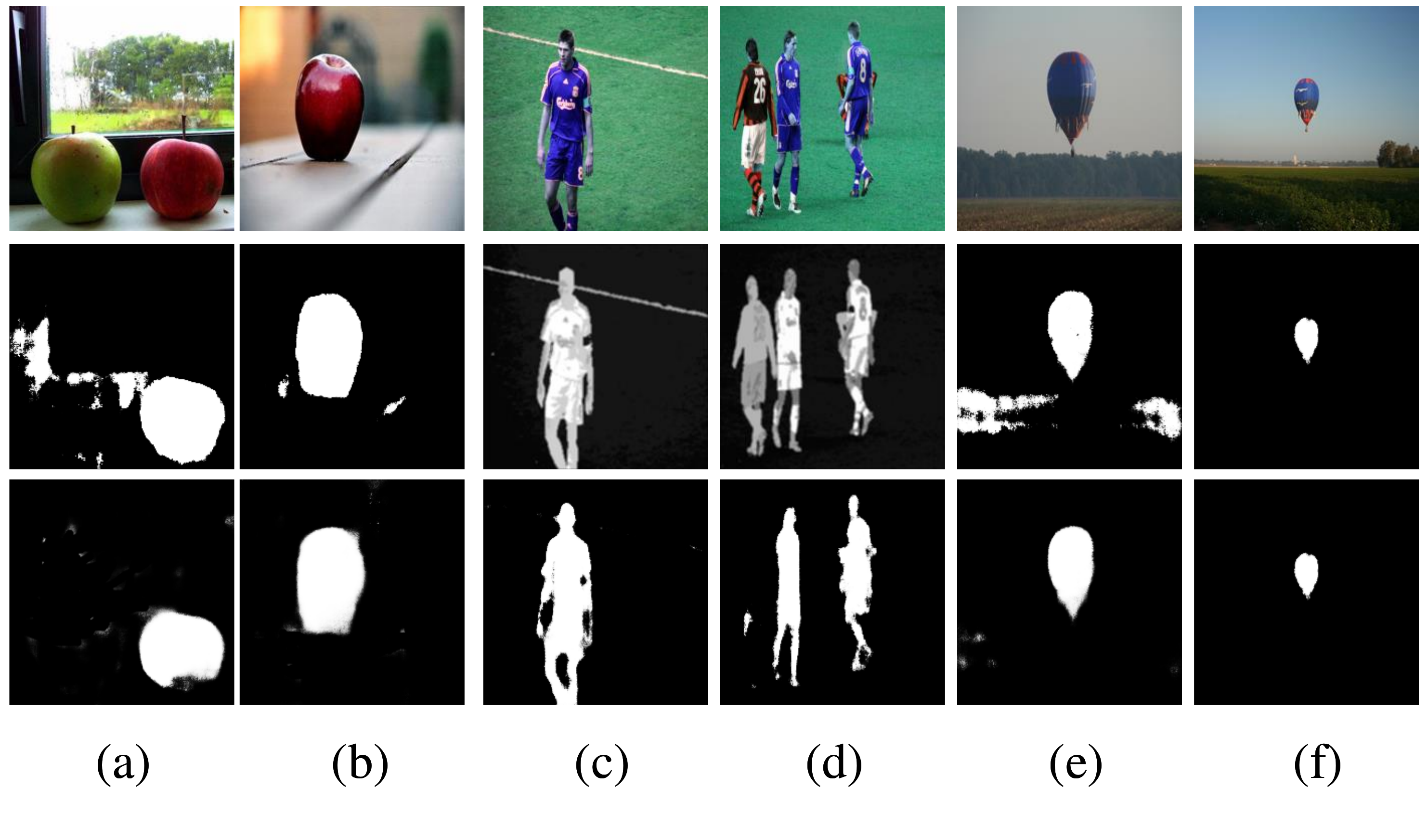}}
	\caption{Results of our CA-FCN model with and without the feature merging strategy. Without feature merging, the model outputs saliency maps (middle row) distracted by other salient regions and backgrounds. CA-FCN produces co-saliency maps of multi-images with only common salient objects by incorporating feature merging (bottom row).}
	\label{fig:merging}
\end{figure}

\textbf{The efficiency of feature merging.} Merging features from each input image is a crucial part of our model. Without the merging step, the model will output saliency maps for each input image. Considering the case that some images within one group may contain salient objects absent in other images. With the merging operation, the undesired salient regions will be suppressed and co-salient objects will be highlighted and stand out from the backgrounds, as shown in Fig.~\ref{fig:merging}. Moreover, the quantitative results will be improved by a considerable margin, w.r.t. $\mathbf{F}_{\beta}$, $\textrm{MAE}$, $\textrm{AUC}$, $\textrm{AP}$, $\mathbf{F}_\beta ^\omega$ and $\mathbf{S}_\alpha$ on the three test datasets, which are tabulated in Table~\ref{Table:merging}.

\begin{table*}[!htb]
	\caption{Quantitative results w.r.t. $\mathbf{F}_{\beta}$, MAE, AUC, AP, $\mathbf{F}_\beta ^\omega$ and $\mathbf{S}_\alpha$ w/o merging features strategy, where w/o and w/ represent without and with merging the individual image features, respectively.}
	\vspace{-0.3cm}
	\begin{center}
		\resizebox{\textwidth}{!}{
		\begin{tabular}{|l|c|c|c|c|c|c||c|c|c|c|c|c||c|c|c|c|c|c|}
			\hline
              \multicolumn{1}{|c|}{\multirow {2}{*}{Methods}} &\multicolumn{6}{c||}{iCoseg~\cite{batra2010icoseg}} &\multicolumn{6}{c||}{MSRC-v2~\cite{winn2005object}} &\multicolumn{6}{c|}{Cosal2015~\cite{zhang2016detection}} \\
			\cline{2-19}
            ~ &$\mathbf{F}_{\beta}$$\uparrow$ &\textrm{MAE}$\downarrow$ &\textrm{AUC}$\uparrow$ &\textrm{AP}$\uparrow$ &$\mathbf{F}_\beta ^\omega$$\uparrow$ &$\mathbf{S}_\alpha$$\uparrow$ &$\mathbf{F}_{\beta}$$\uparrow$ &\textrm{MAE}$\downarrow$ &\textrm{AUC}$\uparrow$ &\textrm{AP}$\uparrow$ &$\mathbf{F}_\beta ^\omega$$\uparrow$ &$\mathbf{S}_\alpha$$\uparrow$ &$\mathbf{F}_{\beta}$$\uparrow$ &\textrm{MAE}$\downarrow$ &\textrm{AUC}$\uparrow$ &\textrm{AP}$\uparrow$ &$\mathbf{F}_\beta ^\omega$$\uparrow$ &$\mathbf{S}_\alpha$$\uparrow$ \\ \hline
            w/o merging &0.7593 &0.1052 &0.9426 &0.8487 &0.7202 &0.7866 &0.8770 &0.0742 &0.9492 &0.9559 &0.8389 &0.8580 &0.6782 &0.1902 &0.9148 &0.8141 &0.5697 &0.7280 \\ \hline
            w/ merging &0.7623 &0.0874 &0.9615 &0.8726 & 0.7374 &0.7977 &0.8832 & 0.0677 &0.9613 &0.9597 &0.8509 &0.8662 &0.6900 &0.1809 &0.9336 &0.8335 & 0.5722 & 0.7310 \\ \hline
            \end{tabular}}
    \label{Table:merging}
	\end{center}
	\vspace{-0.4cm}
\end{table*}

\subsection{Failure examples}
\begin{figure}[htbp]
\centering
	\centerline{\includegraphics[width=0.5\textwidth]{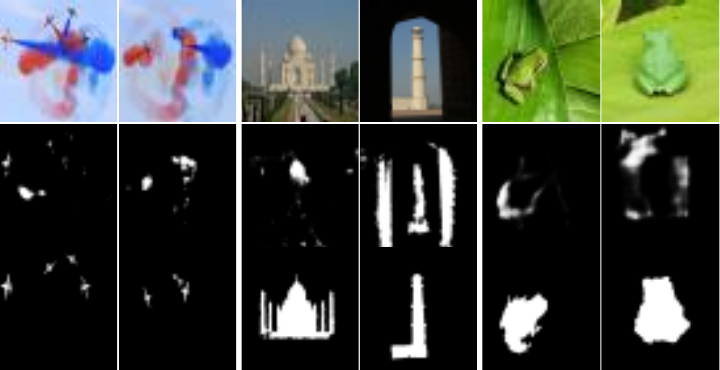}}
	\caption{Failure cases of our approach. Top: input image pairs; middle: co-saliency maps generated by our approach; bottom: ground-truth images.}
	\label{fig:failure}
\end{figure}

Despite the good detection performance on benchmark datasets, our approach also fails on some cluttered cases. Fig.~\ref{fig:failure} shows three pairs of failure examples. For the first failure example, co-salient objects, i.e., airplanes, are very small and high saliency values are assigned to the colourful smoke in the background. For the second example, the focal length of the co-salient objects appearing in the two images are too far apart, which misleads the detection model. For the last example, the co-salient objects have little colour contrast with the backgrounds, which fails our method.

\section{Conclusion}
\label{sec:conclusion}
In this paper, we present a deep supervised co-saliency detection network by integrating a co-attention module into an FCN framework. We apply the co-attention mechanism to the corresponding conv-layers to jointly capture semantic information and retain more additional details to generate accurate co-saliency maps. With the novel co-attention module, the common salient objects of the input image pair are better highlighted while backgrounds and uncommon distractors are suppressed. Extensive experimental results demonstrate that the proposed approach performs favourably on different evaluation metrics against state-of-the-art methods on all adopted benchmark datasets.

For future work, we would address problems such as edge detection, and try to further improve our performance in challenging cases. Also, we are interested in extending our method to videos by leveraging dynamic features and incorporating temporal information. An example solution is \cite{wang2018video}, which extended FCN to video co-saliency by integrating three saliency cues, i,e., inter-video appearance saliency cue, region-contrast saliency cue, and spatial location cue. Besides, how to effectively adapt attention mechanisms to video domains is also crucial. The following works constructed paradigms that may be useful for us. \cite{wang2019revisiting } used attention mechanisms to encode static saliency information and apply LSTMs to learn temporal saliency representation across consecutive video frames. \cite{lai2019video} put forward a composite attention mechanism that learned multi-scale local attentions and global attention priors for enhancing spatio-temporal features.

\bibliographystyle{IEEEtran}
\bibliography{IEEEabrv,references}

\begin{IEEEbiography}[{\includegraphics[width=1in,height=1.25in,clip,keepaspectratio]{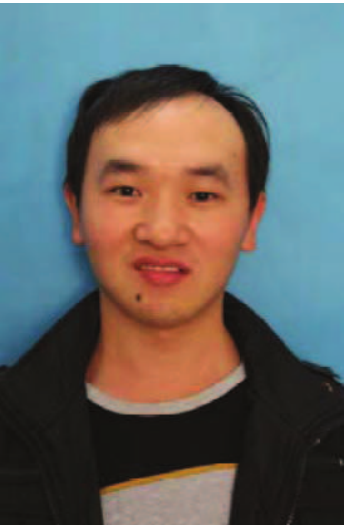}}]{Guangshuai Gao}
received the B.Sc.
degree in applied physics from college of science and the M.Sc. degree in signal and information processing from the School
of Electronic and Information Engineering, from the Zhongyuan
University of Technology, Zhengzhou, China,
in 2014 and 2017, respectively.

He is currently pursuing the Ph.D. degree with the
Laboratory of Intelligent Recognition and Image
Processing, Beijing Key Laboratory of Digital
Media, School of Computer Science and Engineering, Beihang University. His research interests include image processing, remote sensing analysis,
pattern recognition, and digital machine learning.
\end{IEEEbiography}
\begin{IEEEbiography}[{\includegraphics[width=1in,height=1.25in,clip,keepaspectratio]{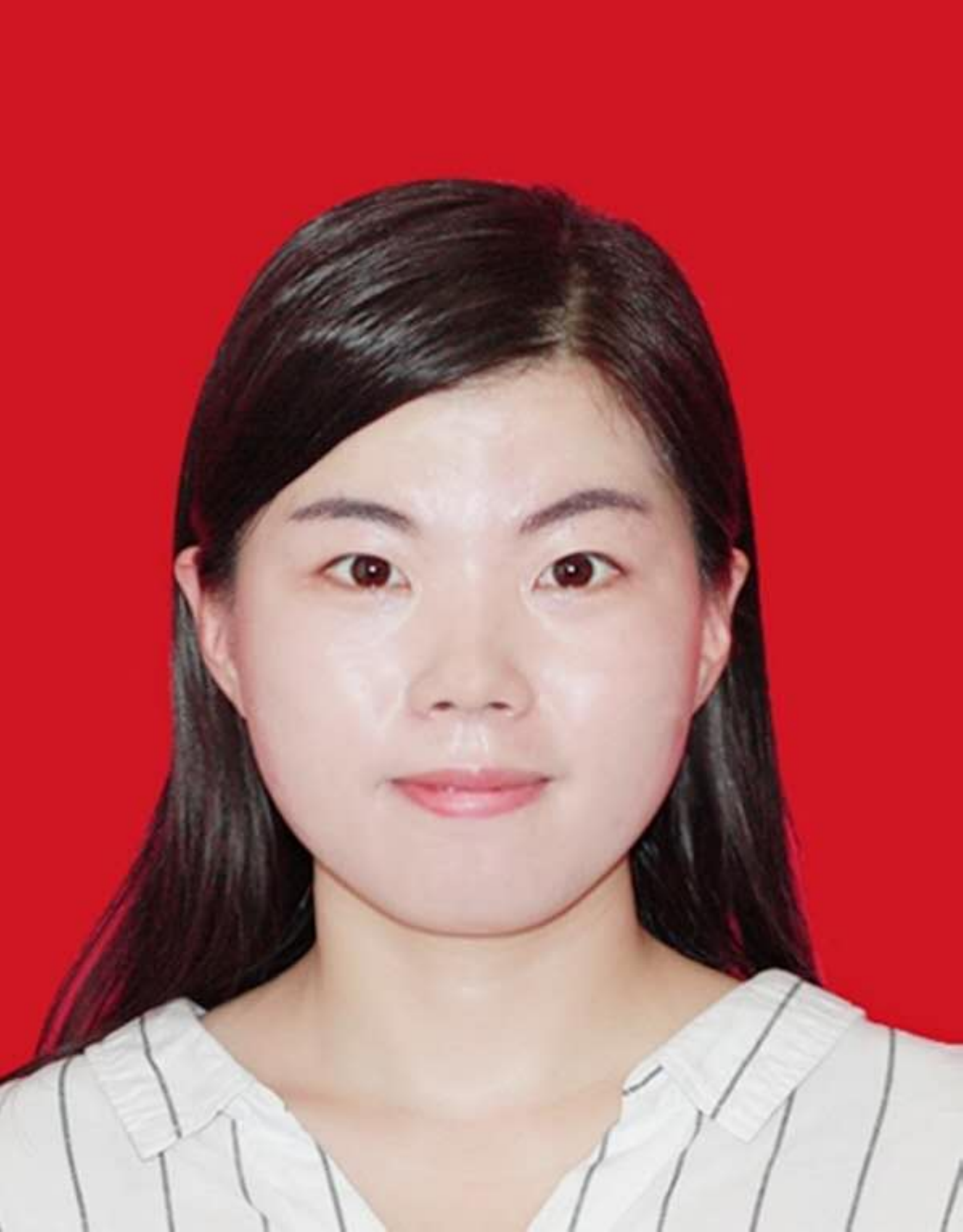}}]{Wenting Zhao}
received the B.Sc. degree from the School
of Computer Science and Engineering, from Xidian University, Xian, China, in 2015, and the M.Sc. degree from the School
of Computer Science and Engineering, from Beihang University, Beijing, China, in 2018.

She is currently working in AI Lab of China Merchants Bank, Shenzhen, China. Her research interests include image processing,
pattern recognition, natural language processing and digital machine learning.
\end{IEEEbiography}

\begin{IEEEbiography}[{\includegraphics[width=1in,height=1.25in,clip,keepaspectratio]{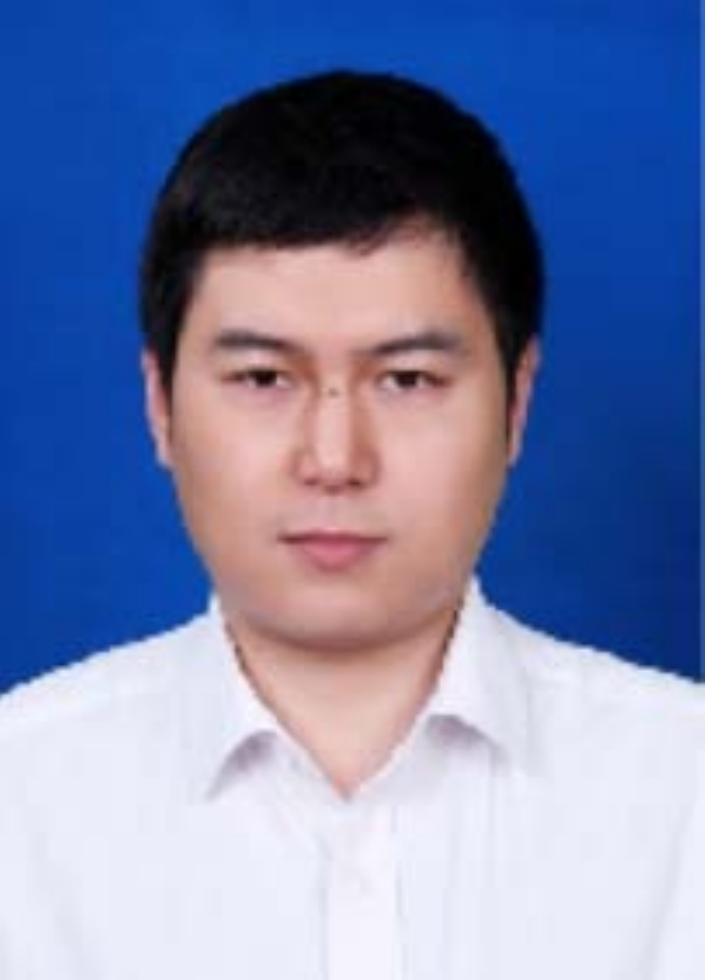}}]{Qingjie Liu}
received the B.S. degree in computer science from Hunan
University, Changsha, China and the Ph.D. degree in computer science from
Beihang University, Beijing, China.

He is currently an Assistant Professor with the School of Computer Science and Engineering, Beihang University.
He is also a Distinguished Research Fellow with the Hangzhou Institute
of Innovation, Beihang University, Hangzhou. His current research interests
include remote sensing image analysis, pattern recognition, and computer
vision. He is a member of the IEEE.
\end{IEEEbiography}

\begin{IEEEbiography}[{\includegraphics[width=1.25in,height=1in,clip,keepaspectratio]{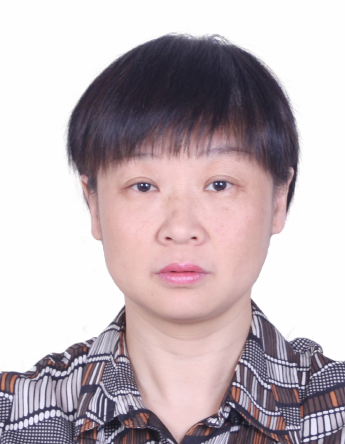}}]{Yunhong Wang}
received the B.S. degree from Northwestern Polytechnical University, Xian, China, in 1989, and the M.S. and Ph.D. degrees from the Nanjing University of Science
and Technology, Nanjing, China, in 1995 and 1998, respectively, all in electronics engineering.

She was with the National Laboratory of Pattern Recognition, Institute of Automation, Chinese Academy of Sciences, Beijing, China, from 1998 to 2004. Since 2004, she has been a Professor with the School of Computer Science and Engineering, Beihang University, Beijing, where she is currently the Director of Laboratory of Intelligent Recognition and Image Processing, Beijing Key Laboratory of Digital Media. Her research results have published at prestigious journals and prominent conferences, such as the IEEE TRANSACTIONS ON PATTERN ANALYS IS AND MACHINE INTELLIGENCE (TPAMI), TRANSACTIONS ON IMAGE PROCESSING (TIP), TRANSACTIONS ON INFORMATION FORENSICS AND SECURITY (TIFS), Computer Vision and Pattern Recognition (CVPR), International Conference on Computer Vision (ICCV), and European Conference on Computer Vision (ECCV). Her research interests include biometrics, pattern recognition, computer vision, data fusion, and image processing. Prof. Yunhong Wang is a Fellow of the IEEE.
\end{IEEEbiography}

\end{document}